\documentclass[runningheads]{llncs}

\usepackage[mobile]{eccv}

\usepackage{eccvabbrv}

\usepackage{graphicx}
\usepackage{booktabs}
\usepackage{multirow}
\usepackage{array}
\usepackage{caption}

\usepackage[accsupp]{axessibility}  

\usepackage{hyperref}

\usepackage{orcidlink}

\usepackage{amssymb}
\usepackage{adjustbox}
\usepackage{booktabs}
\usepackage{pbox}

\begin{document}

\title{MMEarth: Exploring Multi-Modal Pretext Tasks For Geospatial Representation Learning}

\titlerunning{MMEarth: Multi-Modal Pretext Tasks For Representation Learning} 

\author{Vishal Nedungadi\inst{}\orcidlink{0009-0008-3054-7468} \and
Ankit Kariryaa\inst{}\orcidlink{0000-0001-9284-7847} \and
Stefan Oehmcke\inst{}\orcidlink{0000-0002-0240-1559} \and
Serge Belongie\inst{}\orcidlink{0000-0002-0388-5217} \and
Christian Igel\inst{}\orcidlink{0000-0003-2868-0856} \and
Nico Lang\inst{}\orcidlink{0000-0001-8434-027X}}

\authorrunning{V.~Nedungadi et al.}

\institute{University of Copenhagen, Denmark
\\
\email{vishalned@gmail.com, nila@di.ku.dk}}

\maketitle

\begin{abstract}
The volume of unlabelled Earth observation (EO) data is huge, but many important applications lack labelled training data. However, EO data offers the unique opportunity to pair data from different modalities and sensors automatically based on geographic location and time, at virtually no human labor cost.
We seize this opportunity to create \emph{MMEarth}, a diverse multi-modal pretraining dataset at global scale. Using this new corpus of 1.2 million locations, we propose a \textit{Multi-Pretext Masked Autoencoder (MP-MAE)} approach to learn general-purpose representations for optical satellite images. 
Our approach builds on the ConvNeXt V2 architecture, a fully convolutional masked autoencoder (MAE).
Drawing upon a suite of multi-modal pretext tasks, we demonstrate that our MP-MAE approach outperforms both MAEs pretrained on ImageNet and MAEs pretrained on domain-specific satellite images. This is shown on several downstream tasks including image classification and semantic segmentation.
We find that pretraining with multi-modal pretext tasks notably improves the linear probing performance compared to pretraining on optical satellite images only. This also leads to better label efficiency and parameter efficiency which are crucial aspects in global scale applications.
\footnote{The MMEarth dataset is available on the project page: \href{https://vishalned.github.io/mmearth/}{vishalned.github.io/mmearth}. The dataset construction code is available here: \href{https://github.com/vishalned/MMEarth-data}{github.com/vishalned/MMEarth-data}. The MP-MAE code for training and evaluation is available here: \href{https://github.com/vishalned/MMEarth-train}{github.com/vishalned/MMEarth-train}.}

\end{abstract}

\section{Introduction}
Learning representations through self-supervision has seen major advancements in recent years~\cite{balestriero2023cookbook,He_2022_CVPR,oquab2023dinov2,mohamed:22,assran2023self}. This gives hope that even applications with limited supervision data may profit from the promises of the deep learning revolution. However, some self-supervised techniques still fall short when evaluated on image domains other than ImageNet, \eg, ground-level images from the natural world \cite{van2021benchmarking}.
Thus, to develop generalizing representation learning techniques, it is necessary to have suitable large-scale datasets from different domains. 
As an alternative to the ``pure'' self-supervision paradigm, multi-modal data offer great potential for learning good semantic representations \cite{radford2021learning,bachmann2022multimae,mizrahi2023m}.
Aligned multi-modal datasets are key for advancing two major research directions in computer vision: 
i) exploiting multi-modal data for representation learning, and 
ii) advancing representation learning to exploit multi-modal data for inference.

One domain that is particularly well-suited for studying the potential of multi-modal data is Earth Observation (EO). Geolocated data from different sensors can automatically be aligned at virtually no human labor cost. Aligning such EO data at large scale has become substantially easier thanks to cloud-based platforms like Google Earth Engine \cite{gorelick2017google} and Microsoft Planetary Computer \cite{microsoft_open_source_2022_7261897}.
Furthermore, real-world remote sensing applications are often hampered by a lack of high-quality reference data. Human annotation of satellite data is not always possible from the images alone, for example when mapping carbon stocks or species abundance. 
Such applications require field measurements by experts that are costly, time consuming, and do not scale to large areas. Other high-quality measurement systems, such as airborne or spaceborne LiDAR, can provide a means to scale reference data, but still provide a biased sample of the Earth, due to the limits of global sampling.
Here, we present a global dataset called \emph{MMEarth} at the scale of ImageNet-1k \cite{deng2009imagenet} with multi-modal data for 1.2M locations.
We also release our data collection framework for collecting data from different sensors given a list of geospatial coordinates. 
While MMEarth can be used to advance both aforementioned research directions, we demonstrate its potential for the first goal of exploiting multi-modal data for improving representation learning. 
Specifically, our goal is to learn general-purpose representations for optical satellite images from the Sentinel-2 mission that are predictive for a wide range of downstream tasks including crop type, landcover, and climate zone classification.
Thus, in contrast to other work in this area, we neither aim to generalize to other sensor inputs such as Scale-MAE \cite{Reed_2023_ICCV} nor to make use of multiple modalities during inference in downstream tasks such as MultiMAE~\cite{bachmann2022multimae} or 4M~\cite{mizrahi2023m}. 
As shown by previous work, specializing a model to a particular input modality can be beneficial compared to models that aim to generalize to multiple input modalities \cite{mizrahi2023m}. 

To explore the potential of multiple pretext tasks, we build on fully convolutional masked autoencoders (FCMAE) and extend the ConvNeXt V2-MAE approach \cite{woo2023convnext} with multi-modal reconstruction tasks for pretraining. 
As opposed to MAEs based on vision transformers, the FCMAE is well suited for downstream tasks that require different input image sizes, which is inevitable when evaluating the representations on existing EO benchmarks \cite{lacoste2023geobench}. 
Intuitively, such a multi-task pretraining strategy with a shared encoder should lead to representations that generalize better to new downstream tasks that are not known at pretraining time.
We evaluate the generalization potential using Sentinel-2 downstream tasks from GEO-Bench \cite{lacoste2023geobench}, including image-level multi-class and multi-label classification as well as pixel-level semantic segmentation tasks. 
Our contributions can be summarized as follows: 
\begin{enumerate}
    
    \item \textbf{MMEarth dataset} - A global dataset for multi-modal and geospatial representation learning. It consists of 12 modalities including pixel-level and image-level modalities from 1.2 million locations.
    \item \textbf{Method} - We explore the effect of multi-modal pretext tasks for both pixel-level and image-level modalities when learning general purpose representations for interpreting optical images from Sentinel-2. Therefore, we propose a Multi-Pretext Masked Autoencoder (MP-MAE) approach to using MMEarth to learn better representations for Sentinel-2 optical images. 
    \item \textbf{Results} - We show that pretraining MP-MAE on MMEarth outperforms not only MAEs pretrained on ImageNet but also on Sentinel-2 images. While both fine-tuning and linear probing performance are improved, linear probing benefits particularly from the multi-modal pretext tasks.
\end{enumerate}

\section{Related work}

\subsection{Masked image modelling}

Masked Image Modelling (MIM) can see its roots in denoising autoencoders that are trained to reconstruct original images from partially corrupted inputs~\cite{denoising}. This idea was extended to pretrain CNN autoencoders by reconstructing a large area in an image with the aim of learning better representations by conditioning a portion of the image on its surroundings~\cite{pathak2016context}. 
With the success of masked language modelling (for example, BERT~\cite{devlin-etal-2019-bert}) and the rise of Vision Transformers (ViT)~\cite{dosovitskiy2021an},  Masked Autoencoder (MAE) approaches have been developed that reconstruct masked input patches~\cite{dosovitskiy2021an,He_2022_CVPR,xie2022simmim}. When masking a large fraction of the input sequence, MIM is not only a successful pretext task to learn semantic representations, but also leads to a more efficient ViT encoder~\cite{He_2022_CVPR,xie2022simmim}. This masking strategy has also been extended to spatio-temporal modelling to learn from videos \cite{feichtenhofer2022masked, tong2022videomae, wei2022masked,bardes2024revisiting} as well as to learn representations for 3D point clouds~\cite{YuPoint2022}.  
To combine this random masking strategy with the inductive bias of CNNs, we use ConvNeXt V2~\cite{woo2023convnext}, which is a Fully Convolutional MAE (FCMAE). The encoder is implemented with sparse convolutions~\cite{choy20194d} to improve efficiency like in the ViT-based MAE. ConvNeXt V2 also uses a learnable mask token which serves as the input to the convolutional decoder. Instead of using self-attention, the decoder is based on dense convolutions with kernels that cover the full spatial extent of the image. 
This architecture design pretrained on ImageNet yields a high-performing model family of different sizes, ranging from 3.7M (`Atto') to 650M parameters (`Huge'). 
We adapt the ConvNeXt V2 architecture design to satellite images and pretrain on our MMEarth dataset to demonstrate the potential of multi-modal EO data for representation learning.

\subsection{Multi-modal representation learning}

Multi-sensory information has been argued to help humans (and animals) to understand the natural world \cite{deSa1998}, which motivates the development of machine learning approaches that learn representations from multi-modal data.
Successful approaches such as CLIP~\cite{radford2021learning} are based on a contrastive learning objective while learning separate encoders for each input modality. 
This approach has been extended to learn from images, text, and audio extracted from videos~\cite{Argaw2023}.
While contrastive approaches usually rely on aligned samples from various modalities, non-contrastive approaches were proposed to \eg learn image-text representations with a single encoder to avoid the need for paired image-text data~\cite{geng2022multimodal}.
It has been shown that visual tasks are (more or less) related and that it can be more data-efficient to learn multiple tasks jointly~\cite{zamir2018taskonomy}.
In line with these observations, multi-task self-training~\cite{ghiasi2021multitask} leverages pseudo-labels for multi-task learning to yield improved general purpose representations.
In addition to such multi-task objectives, MIM has been extended to the multi-modal image setting~\cite{wei2022masked,bachmann2022multimae,mizrahi2023m}.
Masked feature prediction~\cite{wei2022masked} has been demonstrated as an effective pretext task, \eg by reconstructing the Histograms of Oriented Gradients (HOG) instead of the original image, arguing that the local contrast normalization plays a crucial role.
MultiMAE \cite{bachmann2022multimae} extends the MAE approach~\cite{He_2022_CVPR} to multiple input and output modalities using RGB, semantic label masks, and depth. 
Recently, 4M~\cite{mizrahi2023m} proposed a masked modelling framework to train a single autoencoder for a range of input and output modalities. Instead of reconstructing each modality in the original space (\eg pixels for RGB), each modality is first mapped to discrete tokens using a modality-specific autoencoder. These tokens are then used as the input and output to a transformer-based autoencoder trained with masked modelling, leading to impressive any-to-any reconstruction results. However, their ablation study demonstrates that pretraining with a single input modality can outperform a model trained with multiple input modalities.
While previous works have focused on image modalities that provide pixel-level data, our approach makes use of both pixel-level and image-level modalities.

\subsection{Representation learning in Earth observation}

\subsubsection{Supervised learning at large scale.}
In applications where labelled training data is abundant, supervised learning has been used in a range of applications like tree segmentation~\cite{tucker2023sub}, marine debris detection~\cite{russwurm2023large}, and snow depth estimation~\cite{daudt2023snow}.
However, only a few studies have scaled supervised deep learning to global scales to map land cover~\cite{brown2022dynamic}, canopy height~\cite{lang2023high}, or solar panels~\cite{kruitwagen2021global}.
Additionally, large annotated EO datasets have been created to replace ImageNet with a domain-specific pretraining dataset\cite{christie2018functional,sumbul2019bigearthnet,sumbul2021bigearthnet,wang2022ssl4eo,bastani2023satlaspretrain}. Some bi-modal datasets provide Sentinel-2 optical and Sentinel-1 SAR images in Europe \cite{sumbul2021bigearthnet} and globally around cities~\cite{wang2022ssl4eo}. The benefit of multi-modal input data has been studied for landcover mapping at the scale of Europe in a fully supervised setting~\cite{mommert2023ben}.
In contrast, MMEarth provides a global pretraining dataset with 12 modalities balanced across 14 biomes~\cite{Dinerstein2017}.

\subsubsection{Self-supervised learning.}
Self-supervised approaches developed on ground-level computer vision benchmarks (like ImageNet) can be applied to satellite images~\cite{wang2022ssl4eo,TOLAN2024113888}.
However, these approaches might be suboptimal due to the differences in ground-level and satellite images. Therefore, the EO community has started to explore the unique opportunities offered by EO data. 
We consider three major aspects of EO data for self-supervised representation learning: \emph{geolocation}, \emph{time}, and \emph{modality}.

\textbf{Geolocation.} In an approach called geography-aware self-supervised learning (GASSL)~\cite{ayush2021geography}, the contrastive learning objective leverages the spatio-temporal structure of EO data. Using spatially aligned images, positive image pairs can be created by using observations at different times, avoiding the need for artificial augmentations. In addition, localization is used as a pretext task by predicting the clusters of sampled locations.
SatCLIP~\cite{klemmer2023satclip} focuses on learning location embeddings in a contrastive manner by learning a Sentinel-2 encoder and a location encoder with a contrastive objective.
\textbf{Time.} Seasonal contrast (SeCo~\cite{manas2021seasonal}) is another temporal contrastive approach that learns multiple embedding subspaces that are invariant to either seasonal changes, synthetic augmentations, or all transformations.
Alternatively, image time series can be exploited in masked modelling as was demonstrated in SatMAE~\cite{cong2022satmae}. Presto~\cite{tseng2023lightweight} is a masked modelling approach that focuses on learning representations for multi-modal time series using a light-weight transformer architecture that only relies on pixel time series inputs and ignores textural features.
\textbf{Modality.} Scale-MAE~\cite{Reed_2023_ICCV} aims to generalize to different input resolutions by conditioning on the ground sampling distance (GSD, size of a pixel on the ground).
Most related to our work is SatlasPretrain~\cite{bastani2023satlaspretrain}, a large multi-task dataset for optical satellite images including Sentinel-2. Their goal was \textit{``to label everything that is visible in a satellite image''} using various data sources (Open Street Map, lidar scans, landcover maps) and human annotators (domain experts and Mechanical Turk workers). 
Our MMEarth dataset creation differs, as it is fully automatic without relying on human annotators and also considers modalities that are not \emph{explicitly} \emph{visible to humans} in optical images. Such modalities include everything that cannot be annotated by a human like radar, temperature, precipitation, or canopy height. We hypothesize that predicting such cross-modal relations from images requires the extraction of semantic features.

All these works provide important insights into how to make better use of satellite data and its metadata in representation learning.
Our work extends along the last aspect --- exploiting multiple modalities --- whereas space and time serve as the underlying structure for pairing data from different sources automatically without relying on human annotation. Furthermore, we treat geolocation and time as two additional modalities in our framework.

\section{MMEarth dataset}

\begin{figure}[tb]
  \begin{subfigure}{0.615\linewidth}
    \includegraphics[width=\linewidth]{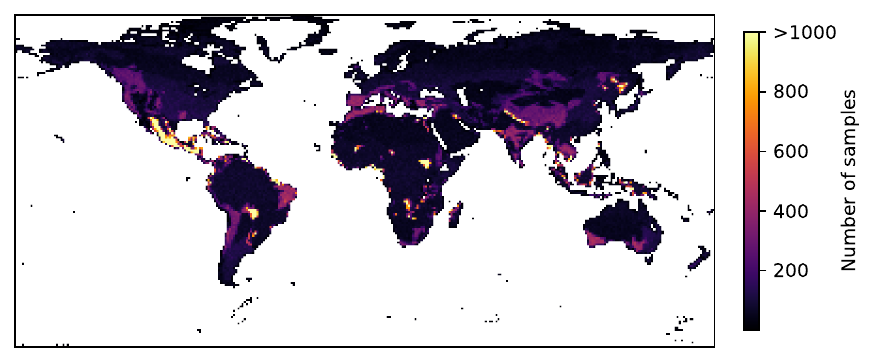}
    \caption{Spatial distribution.}
    \label{fig:distribution_spatial}
  \end{subfigure}
  \hfill
  \begin{subfigure}{0.36\linewidth}
    \includegraphics[width=\linewidth]{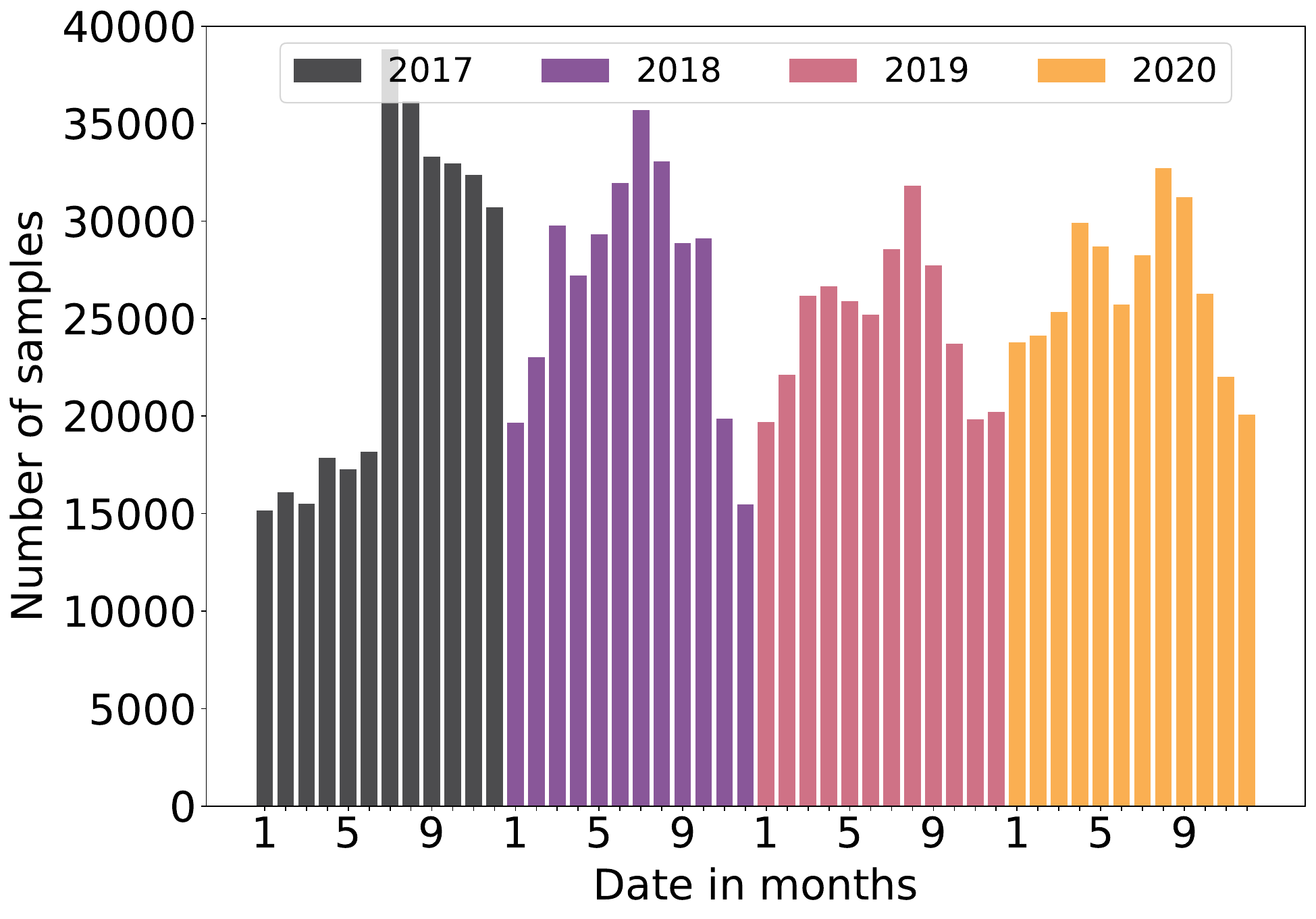}
    \caption{Temporal distribution.}
    \label{fig:distribution_temporal}
  \end{subfigure}
  \caption{\textbf{MMEarth dataset coverage.} With a balanced sampling scheme across 14 biomes and 4 years, we collected aligned multi-modal data from 12 modalities using the Google Earth Engine platform \cite{gorelick2017google} at 1.2M locations.}
  \label{fig:data_distribution}
\end{figure}

\begin{table}[htb]

\caption{\textbf{Modalities in the MMEarth dataset.} Each location provides 12 aligned modalities with a total of 46 bands. The upper half of the modalities provide dense data at the pixel level and the bottom half at the image level.  }
\label{tab:dataset_modalities}
\adjustbox{max width=\textwidth}{
\begin{tabular}{@{}llllcl@{}}
\toprule
\hspace{0.5cm}                      & Sensor/Product     & Description         & Data type      & \#Bands & Band information                                   \\ \midrule
\multirow{6}{*}{\rotatebox[origin=c]{90}{Pixel-level}} & Sentinel-2         & Optical   & continuous  & 12 & multi-spectral B1-B12 for L1C/L2A product                         \\
                                        & Sentinel-1         & SAR                 & continuous  & 8     & VV, VH, HV, HH for ascending/descending orbit      \\
                                        & Aster DEM          & Elevation           & continuous  & 2     & elevation, slope                                   \\
                                        & ETH-GCHM           & Vegetation height   & continuous  & 2     & canopy height, uncertainty (STD)                  \\
                                        & Dynamic World      & Landcover           & categorical & 1     & 9 landcover categories                            \\
                                        & ESA World Cover    & Landcover           & categorical & 1     & 11 landcover categories                            \\ \midrule 
\multirow{6}{*}{\rotatebox[origin=c]{90}{Image-level}} & Biome              & Landcover           & categorical & 1     & 14 terrestrial ecosystem categories                            \\
                                        & Ecoregion          & Landcover           & categorical & 1     & 846 ecoregion categories                           \\
                                        & ERA5 temperature   & Climate reanalysis  & continuous  & 9     & mean, min, and max of [year, month, previous month]  \\
                                        & ERA5 precipitation & Climate reanalysis  & continuous  & 3     & total precipitation of [year, month, previous month] \\
                                        & Geolocation        & Latitude, Longitude & continuous  & 4     & cyclic encoding of latitude and longitude          \\
                                        & Date (Sentinel-2)  & Month of the year   & continuous  & 2     & cyclic encoding of the month               \\ \bottomrule
\end{tabular}
}
\end{table}

MMEarth contains data for 1.2 million locations distributed around the world, making its optical image count comparable to that of ImageNet-1K~\cite{deng2009imagenet}. At each location, data from 12 geo-aligned modalities were collected, grouped into pixel-level and image-level modalities as per Table~\ref{tab:dataset_modalities}. 
The six pixel-level modalities represent raster data of size $128\times128$ pixels which capture $1.28\times1.28$~km on the ground (\eg, Sentinel-2, Sentinel-1, Aster DEM, Dynamic World, and ESA World Cover). 
The remaining six image-level modalities represent scalar values for each location (\eg, Biome, Ecoregion, ERA5 temperature, ERA5 precipitation, Geolocation, and Sentinel-2 observation date). 
The size of the input images is large enough for typical downstream EO tasks ~\cite{lacoste2023geobench} and yields a reasonable dataset volume.
MMEarth was sampled uniformly across 14 biomes (Fig.~\ref{fig:data_distribution}, \eg, Mangroves, Temperate Conifer Forests, Tundra, etc.)~\cite{Dinerstein2017}. 
To increase diversity, we considered data from the four years 2017--2020 (Fig.~\ref{fig:data_distribution}).
Further, we ensured that time-critical modalities were collected around the Sentinel-2 observation date, which serves as the reference. 
Additionally, all pixel-level modalities were re-projected to the Sentinel-2 10m grid, if needed, yielding a harmonized data cube.
In addition to the full dataset, we provide two subsets to facilitate research in multi-modal representation learning with limited compute resources: \emph{MMEarth100k} (100k locations at $128\times128$ pixels) and \emph{MMEarth64} (1.2M locations center-cropped to $64\times64$ pixels); see supplementary material for details.
All the data were downloaded from Google Earth Engine (GEE)~\cite{gorelick2017google}.
In the following subsections, we provide a brief overview of the pixel-level and image-level modalities listed in Table~\ref{tab:dataset_modalities}. More details are provided in the supplementary material.

\subsection{Pixel-level modalities}

\textbf{Sentinel-2.} 
An optical satellite mission that provides global coverage at least every 5 days at 10m resolution. Its multi-spectral sensor comprises of 12 bands, with four bands RGBN (RGB and near infrared) at 10m and near and short-wave infrared bands at 20m and 60m resolution, respectively. We upsample all bands to the 10m grid.
Sentinel-2 provides two processing levels: top of atmosphere reflectance (L1C) and bottom of atmosphere reflectance (L2A). L2A is atmospherically corrected, harmonizing the spectral data, but is not yet available for all years on a global scale. To ensure global coverage for the years 2017--2020, MMEarth includes both L2A and L1C data.
\textbf{Sentinel-1.} Synthetic Aperture Radar (SAR) data at 10m resolution, collected in two viewing directions along an ascending and descending orbit, resulting in four bands each. 
\textbf{Aster GDEM.} A Global Digital Elevation Model derived from the Advanced Spaceborne Thermal Emission and Reflection Radiometer (ASTER). The Aster GDEM data covers 99\% of the Earth's landmass at 15m resolution. We include elevation and slope.
\textbf{ETH-GCHM.} A 10m Global Canopy Top Height Map~\cite{lang2023high} for the year 2020 derived from Sentinel-2 and sparse vertical structure data from GEDI~\cite{dubayah2020global}, a spaceborne LiDAR mission. 
\textbf{Dynamic World.} A land cover dataset that contains semantic segmentation of Sentinel-2 images for nine categories at 10m resolution. We aggregate the per-image maps for each year to reduce noise in the data.
\textbf{ESA World Cover.} A 10m global land cover map based on Sentinel-1 and Sentinel-2 data with 11 categories for the year 2020.

\subsection{Image-level modalities}
\textbf{Biome.} A biome is a geographical region with specific vegetation, climate and animal life. 
We use the RESOLVE Ecoregions dataset~\cite{Dinerstein2017} that consists of 14 terrestrial biomes. 
\textbf{Ecoregion.} The RESOLVE ecoregions~\cite{Dinerstein2017} include 846 terrestrial ecoregions. These are a more fine-grained classification than biomes. For example, the ecoregions \emph{Central African Mangroves} and \emph{Indochina Mangroves} belong to the biome \emph{Mangroves}.
\textbf{ERA5 Temperature and Precipitation.}  ERA5 reanalysis data provides climate information back to 1950.
Based on the Sentinel-2 observation date, we collect the corresponding mean, min, and max temperature as well as the total precipitation for the month, previous month, and year. 
\textbf{Geolocation.} We use cyclic encodings for the latitude and longitude of each image center. This ensures that for the longitude the cyclic property is maintained (transition from West to East).
\textbf{Date.} We use a cyclic encoding of the month of the Sentinel-2 observation date as a proxy for the season.

\section{Methodology}

\subsection{Multi-Pretext Masked Autoencoder}

\begin{figure}[tb]
  \centering
    \includegraphics[width=\textwidth]{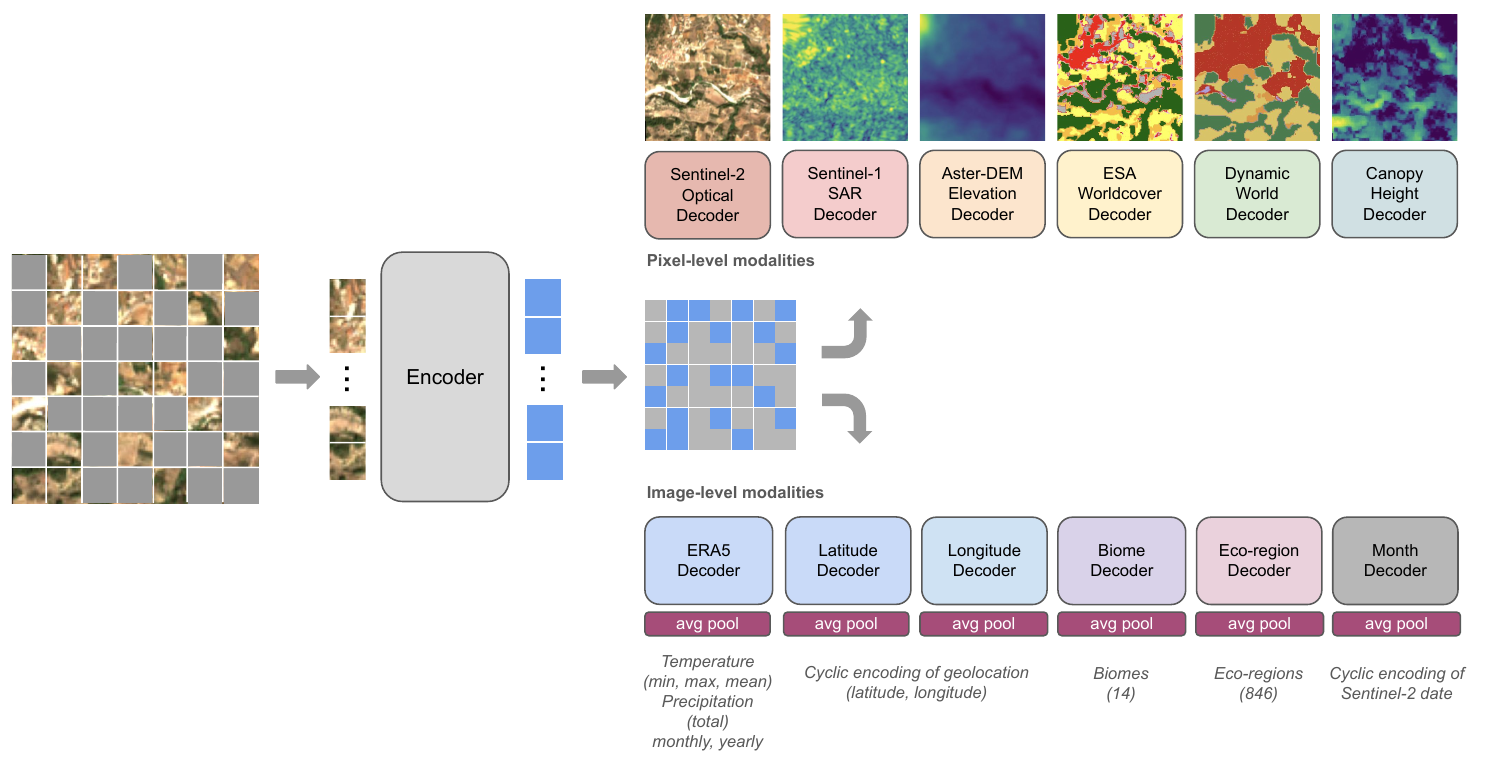}
  \caption{\textbf{Multi-Pretext Masked Autoencoder (MP-MAE).} Our approach extends Masked Autoencoders, which reconstruct only the input image, by incorporating multiple pretext tasks using aligned pixel-level as well as image-level modalities.}
  \label{fig:method}
\end{figure}

In this section, we describe the proposed Multi-Pretext Masked Autoencoder (MP-MAE) approach as illustrated in Fig.~\ref{fig:method}. It builds on the promising results of MIM with the ConvNeXt V2 architecture.

\subsubsection{Encoder.}
The input image $x$ is divided into patches (or tokens) on which a random mask is applied to yield a masked image $x'$.
The encoder $f_{\theta}$ takes the visible patches as the input and returns an encoding $z$ for every visible patch. Here, the ConvNeXt V2~\cite{woo2023convnext} architecture serves as the backbone which is implemented with sparse convolutions~\cite{choy20194d} to improve efficiency.
We investigated two key modifications to adapt the ConvNeXt V2 architecture to medium-resolution satellite images.
First, we adapted the patch size. The original ConvNeXt V2-MAE has been developed for pretraining on ImageNet and assumes input images of size $224\times224$ pixels, which are split into  $7\times7$ patches of size $32\times32$ pixels. 
MMEarth images are of size $128\times128$ pixels, and we randomly crop them to $112\times112$ pixels when feeding them into the MP-MAE encoder, which has a reduced  patch size of $16\times 16$ pixels to preserve the $7\times7$ patch layout. This adjustment is crucial as the number of patches is coupled with the optimal masking ratio. 
Second, we avoid early downsampling by modifying the first layer. In the original ConvNeXt V2 encoder, the first layer is a learned convolutional downsampling layer, which yields feature maps at 1/4 of the input resolution. While this approach might be efficient and loses only little spatial information that is not needed for ImageNet classification, it may not be optimal 
for medium-resolution satellite images such as Sentinel-2 and for pixel-level downstream tasks such as semantic segmentation with U-Net architectures~\cite{ronneberger2015u}, where high-resolution features can inform same-resolution representations via the skip connections. 
Therefore, we replace the first layer with an initial convolutional layer with kernel size 3 and stride 1 to learn feature maps at the input resolution followed by a depth-wise convolutional downsampling layer.

\subsubsection{Decoders.}
We follow the ConvNeXt V2-MAE~\cite{woo2023convnext} decoder design and use shallow decoders with one block.  
We apply the same random mask to all pixel-level modalities and learn a separate decoder $h_{t}$ for each pretext task. 
We consider $T$ pretext tasks and treat most modalities as individual tasks, but group the climate variables and split latitude and longitude (see Fig.~\ref{fig:method}).
To reconstruct the masked patches, a learnable mask token is used as a placeholder when combining the embedding tokens to get a dense 2D-input for the decoders. Therefore, the decoders are based on standard convolutions.
For the image-level modalities, we introduce a global average pooling over the decoded non-visible patches before the final linear prediction layer. Hence, we follow the same strategy as for the pixel-level modalities, wherein the encoder is encouraged to learn representations that provide the semantic context that is needed to decode the masked tokens.

\subsection{Multi-task loss}

All pretext task targets $y_{t}$ are reconstructed by applying the same random mask to prevent the model from learning any shortcuts between the input and the targets. We follow prior work and compute the losses only over the non-visible patches.
We experimented with two settings for the multi-task loss function~\cite{vandenhende2021multi,kendall2018multi}
\begin{equation}
    \mathcal{L} = \sum_{t=1}^{T} \frac{1}{\hat{\sigma}_{t}^{2}}\mathcal{L}_t\left(h_{t}(f_{\theta}(x)), y_{t}\right) + \log \hat{\sigma}_{t} ,
\end{equation}
with task-specific loss $\mathcal{L}_t$ (\eg, cross entropy for classification and mean squared error for regression) and variance $\hat{\sigma}_{t}^{2}$.
First, we applied equal weighting with $\hat{\sigma}_{t}^{2}=1$ for all decoders, followed by task-uncertainty weighting where $\hat{\sigma}_{t}^{2}$ is learnable for each decoder~\cite{kendall2018multi,bachmann2022multimae}.
The second setting was expected to aid in handling noisy targets, as task-specific weights naturally decrease for such pretext tasks~\cite{kendall2018multi}.

\section{Experimental setup}

In our problem setup, the downstream tasks are not known at pretraining time. Given a pretrained encoder, we follow standard practices and evaluate fine-tuning (FT) and linear probing (LP) performance on new downstream tasks. While both approaches learn the last linear layer of a neural network from scratch, FT updates the pretrained encoder and LP uses a frozen encoder.
Implementation details are provided in the supplementary material. 

\subsection{Evaluation on downstream tasks}
For downstream tasks, we considered all five Sentinel-2 datasets provided in the GEO-Bench~\cite{lacoste2023geobench} EO dataset collection\footnote{We use the latest GEO-Bench version v1.0 in which datasets were class-balanced.}. 
These datasets provide a diverse set of tasks and geographical coverage (Global, Europe, Africa).
GEO-Bench provides harmonized subsets of the original datasets with comparable training set sizes and balanced class distributions. For clarity, we refer to the datasets with the number of training samples to avoid confusion with prior work using the original datasets.
We summarize the characteristics of these five datasets in Table~\ref{table:geobench-downstream} and provide more details in the following sections. 

\begin{table}[tbp]
\caption{\textbf{Downstream task datasets.} We evaluated on five different datasets which involve image-level classification (multi-label, multi-class) and pixel-level semantic segmentation. These datasets are modified versions from the GEO-Bench benchmark~\cite{lacoste2023geobench}.}
\label{table:geobench-downstream}
\adjustbox{max width=\textwidth}{
\begin{tabular}{@{}lllccc@{}}
\toprule
Dataset Name                             & Description                               & Task         & Categories & Input [px]         & Train/Val/Test \\ \midrule
BigEarth20k~\cite{sumbul2019bigearthnet} & Landcover classification in Europe        & Multi-label  & 43         & 120$\times$120 & 20k/1k/1k      \\
So2Sat20k~\cite{zhu2020so2sat}           & Global local climate zones classification & Multi-class  & 17         & 32$\times$32   & 20k/1k/1k      \\
EuroSat2k~\cite{helber2019eurosat}      & Landcover classification in Europe        & Multi-class  & 10         & 64$\times$64   & 2k/1k/1k      \\ 
\midrule
Cashew1k~\cite{Yin2023}                  & Cashew plantations in Benin (Africa)      & Segmentation & 7          & 256$\times$256 & 1.3k/400/50    \\
SAcrop3k~\cite{SAcrop}                   & Crop-type segmentation in South Africa    & Segmentation & 10         & 256$\times$256 & 3k/1k/1k       \\ \bottomrule
\end{tabular}
}
\end{table}

\noindent\textbf{Image classification:}
\textbf{BigEarthNet}~\cite{sumbul2019bigearthnet} is a multi-label land cover classification problem with 43 categories and data from 10 countries in Europe. While the original dataset contains a total of 590k Sentinel-2 image samples, the GEO-Bench version, referred to as `BigEarth20k', contains 20k training samples. 
\textbf{So2Sat}~\cite{zhu2020so2sat} is a multi-class land cover classification task with 17 categories that classifies local climate zones. The data is collected globally from urban areas and the original data contains 400k pairs of Sentinel-2 and Sentinel-1 image samples. The GEO-Bench version contains 20k training samples and is referred to as `So2Sat20k'. 
\textbf{EuroSat}~\cite{helber2019eurosat} is a multi-class land use and land cover classification dataset covering Europe with 10 categories. The original dataset contains 27k image samples, and the GEO-Bench version `EuroSat2k' has roughly 2k training samples. 

\noindent\textbf{Semantic segmentation:}
\textbf{Cashew Plantation}~\cite{Yin2023} is a semantic segmentation task with 7 categories that maps cashew plantations in Benin, Africa. The `Cashew1k' consists of 1k training samples. 
\textbf{SA-Crop-Type}~\cite{SAcrop} is another semantic segmentation problem with 10 categories that maps crop type in South Africa. The `SAcrop3k' consists of 3k training samples. 

\noindent\textbf{U-Net evaluation for semantic segmentation:}
We adopt a U-Net~\cite{ronneberger2015u} architecture to evaluate the pretrained encoders. 
These U-Nets are fine-tuned in two phases, referred to as FT in the results. First, we trained the randomly initialized U-Net decoder with the frozen pretrained encoder for 50 epochs, and then fine-tuned the full model for another 150 epochs.

\noindent\textbf{Evaluation metrics:} Following GEO-Bench~\cite{lacoste2023geobench}, we use the micro-averaged overall accuracy (Acc.) for multi-class, the micro-averaged F1-score (F1) for multi-label classification, and the macro-averaged intersection over union (IoU) for semantic segmentation tasks.

\section{Results}

We investigated two research questions: \textit{How can domain specific pretraining improve representations?} and \textit{How do multi-modal pretext tasks impact representations?}

\subsection{How can domain specific pretraining improve representations?}

\noindent\textbf{Switching to domain-specific data is not enough.} 
The results from Table~\ref{tab:results_domain_specific} indicate that just replacing the ImageNet pretraining data with domain-specific optical satellite images does not improve downstream performance.
We propose slightly adjusting the encoder design and the patch size to adapt to medium resolution satellite images like Sentinel-2 (10m pixels on the ground).
Reducing the patch size from 32 to 16 or 8 pixels, for image sizes of 112 or 56 pixels, respectively, is crucial to adapt to the smaller image size of the medium resolution satellite images. This preserves the number of patches to be 7$\times$7 as in the original ConvNeXt V2 ImageNet setting and is related to an optimal masking ratio \cite{xie2022simmim}. 
With these modifications, domain-specific pretraining improved fine-tuning and linear probing performance across all datasets compared to pretraining on ImageNet. The improvement is smallest for FT on Cashew1k with 3pp and largest for FT on So2Sat with 12pp or 17pp, when training on image sizes of 112 and 56, respectively. 

\noindent\textbf{Multi-spectral images are beneficial.}
Since Sentinel-2 images provide 12 bands in total, we studied the impact of pretraining on multi-spectral images and found that this can improve downstream performance depending on the tasks (see Table~\ref{tab:results_domain_specific}). 
So2Sat20k is an exception where the model using only the RGB channels achieved high performance. 
The following multi-pretext results are based on using all Sentinel-2 bands as input.

\begin{table}[tbp]
\centering
\caption{\textbf{Domain specific pretraining on optical images.} Fine-tuning (FT) and linear probing (LP) performance using MAE pretrained on ImageNet (RGB), Sentinel-2 RGB (MMEarth64-S2rgb, MMEarth-S2rgb), and Sentinel-2 multi-spectral (MMEarth64-S2). Lowering the patch size is crucial to adapt to the smaller medium resolution satellite images. 
Using all 12 bands improves accuracy.
Domain-specific pretraining improves both FT and LP results.}
\label{tab:results_domain_specific}
\adjustbox{max width=\textwidth}{
\begin{tabular}{lrrcccc}
\toprule
Pretrain data & \begin{tabular}[l]{@{}c@{}}Image \\size\end{tabular} & \begin{tabular}[l]{@{}c@{}}Patch \\size\end{tabular} & \begin{tabular}[l]{@{}c@{}}BigEarth20k\\(F1$\uparrow$) \\ FT/LP \end{tabular} & \begin{tabular}[c]{@{}c@{}}So2Sat20k\\(Acc.$\uparrow$)\\ FT/LP \end{tabular} & \begin{tabular}[c]{@{}c@{}}Cashew1k\\(IoU$\uparrow$)\\ FT\end{tabular} & \begin{tabular}[c]{@{}c@{}}SAcrop3k\\(IoU$\uparrow$)\\ FT\end{tabular} \\ \midrule
ImageNet &    224 &  32   &   55.7/25.9   & 36.6/24.0 &      77.1 &      26.7 \\
MMEarth-S2rgb &  128 & 32 &   55.5/29.8 & 38.1/22.7 &      77.2 &      23.8 \\
MMEarth-S2rgb &  112 & 16 &   62.3/33.9 & 48.8/29.8 &      79.3 &      28.7 \\
MMEarth64-S2rgb & 56   & 8 & 64.9/34.4  & \textbf{54.1}/32.5  & \textbf{80.0}  & 33.1  \\
MMEarth64-S2    & 56   & 8 & \textbf{65.1}/\textbf{36.2}  & 48.7/\textbf{33.5}  & 79.9  & \textbf{36.0}  \\ \bottomrule
\end{tabular}
}
\end{table}

\subsection{How do multi-modal pretext tasks impact representations?}

\noindent\textbf{Multi-modal pretext tasks improve fine-tuning and linear probing.} Using multi-pretext tasks improved both the FT and LP performance for all downstream tasks (see Table~\ref{tab:results_multi_modal}).
While MAE are known for good FT performance, the gap with the LP performance remains large~\cite{He_2022_CVPR,woo2023convnext}. The representations learned from our multi-pretext objective notably improved LP performance and provide a mechanism towards closing this performance gap. We observed an improvement of $\approx$5pp on BigEarth20k and 8pp on So2Sat20k with respect to pretraining on Sentinel-2 only. 
Here, we report results obtained with the task uncertainty loss. We found that both loss settings work equally well on these downstream tasks (see supplementary). However, the learned task uncertainties provide interesting insights, \eg  the uncertainty for the latitude (N-S) being lower than for the longitude (E-W), which could be explained by the rainfall and vegetation gradients along the North-South axis shaping the appearance of landscapes.

\noindent\textbf{Pixel-level and image-level pretext tasks are useful.}
To study the role of the two groups of pretext tasks, we ran experiments where the Sentinel-2 reconstruction task was extended with only one group (see Table~\ref{tab:results_multi_modal}).
For LP we found that combining the tasks from both groups leads to the best performance on both datasets. For BigEarth20k using only one group did not improve LP performance over pretraining on Sentinel-2 only. For So2Sat20k image-level pretext tasks lead to slightly better LP performance, but pixel-level pretext tasks lead to better FT performance.
While we had expected that pixel-level pretext tasks might be more beneficial for segmentation tasks, this was not confirmed. 
From these experiments we hypothesize that the semantics of the pretext tasks are more important than whether the tasks are pixel- or image-level. 

\begin{figure}[tb]
  \centering
  \begin{subfigure}{0.3\linewidth}
    \includegraphics[width=\linewidth]{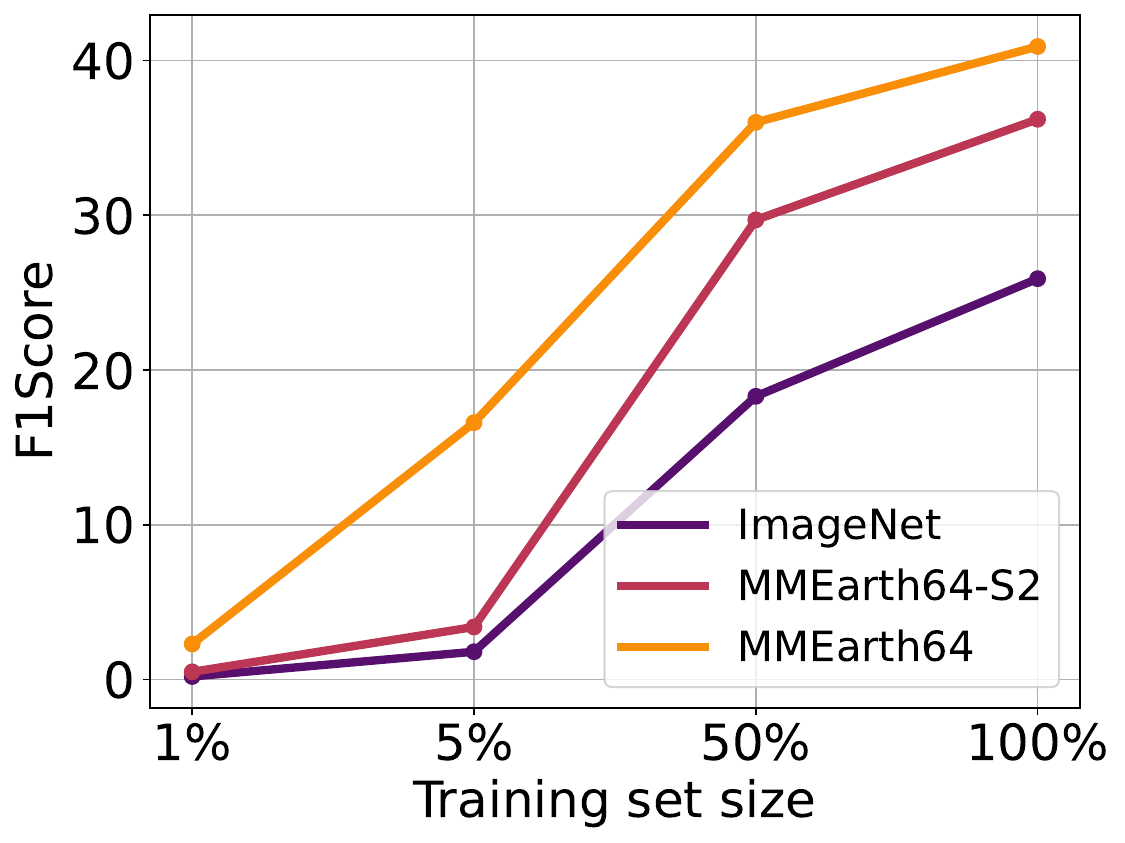}
    \caption{BigEarth20k}
    \label{fig:label_efficient-a}
  \end{subfigure}
  \hfill
  \begin{subfigure}{0.3\linewidth}
    \includegraphics[width=\linewidth]{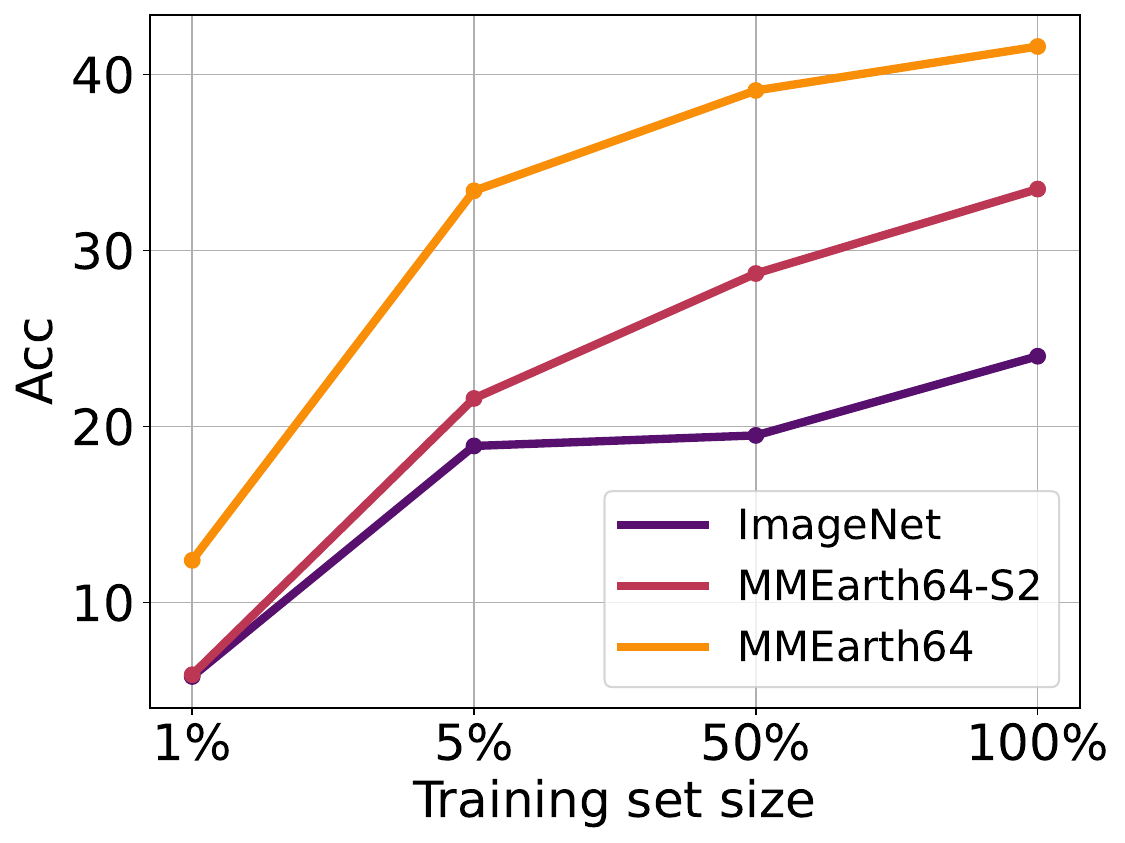}
    \caption{So2Sat20k}
    \label{fig:label_efficient-b}
  \end{subfigure}
  \hfill
  \begin{subfigure}{0.3\linewidth}
    \includegraphics[width=\linewidth]{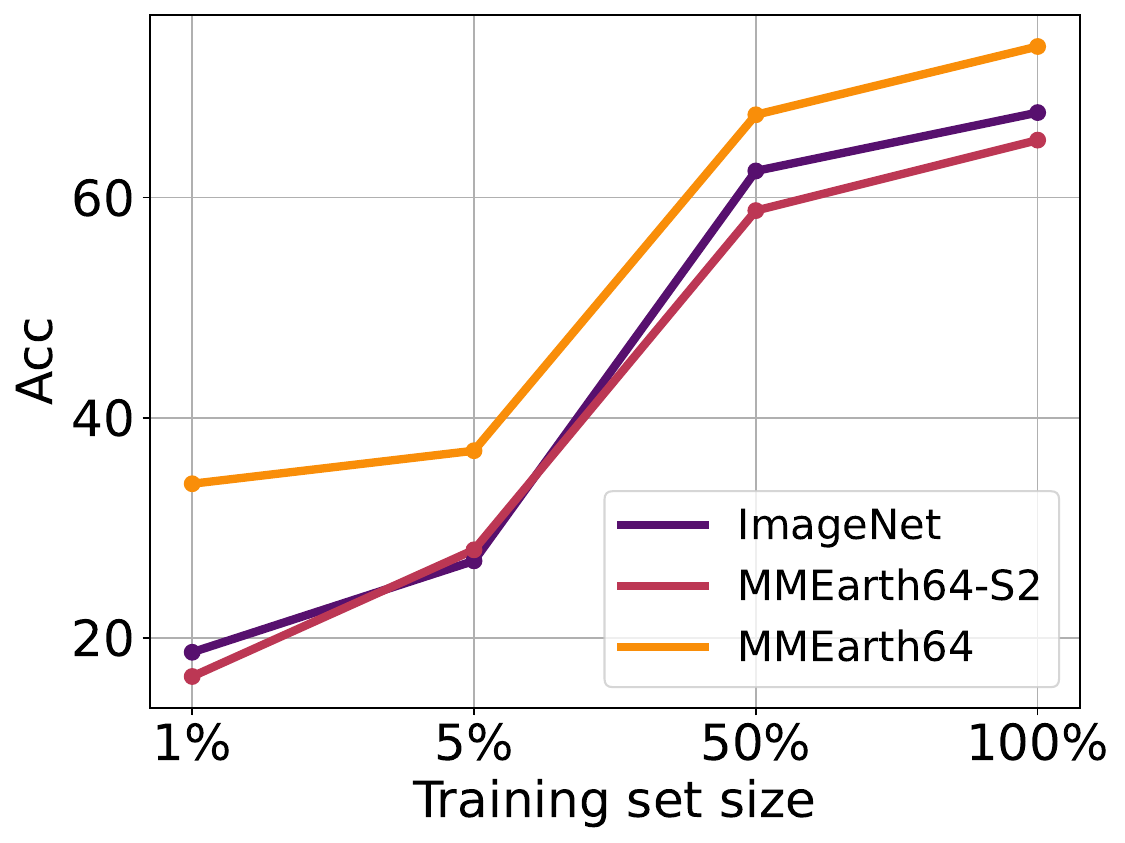}
    \caption{EuroSat2k}
    \label{fig:label_efficient-c}
  \end{subfigure}
  \caption{\textbf{Label efficiency for few-shot downstream performance.} Linear probing performance for varying downstream dataset sizes. MP-MAE (`Atto') pretrained on ImageNet, MMEarth64-S2 (multi-spectral only), MMEarth64 (all multi-modal pretext tasks). }
  \label{fig:label_efficient}
\end{figure}

\noindent\textbf{Multi-modal pretext tasks improve label efficiency.}
Real-world applications often lack large amounts of labelled data. To explore the benefits of pretraining in a limited label regime, we evaluated the LP performance using stratified subsets of the fine-tuning datasets (GEO-Bench partitions). In the most extreme few-shot setting for BigEarth20k, there were only 200 training samples, \ie 4--5 samples/category. For EuroSat2k the extreme few-shot setting has only 20 training samples with 2 samples/category.
Results in Fig.~\ref{fig:label_efficient} show that models pretrained with multi-modal pretext tasks on MMEarth outperformed models trained on only Sentinel-2 or only ImageNet in few-shot linear probing.
While EuroSat is an easy task for supervised learning, GEO-Bench provides a rather small subset, which makes it a challenging few-shot task. For EuroSat2k, we found that general image features learned on ImageNet are a strong baseline working on par with our model trained on Sentinel-2 images only. Even in the extreme case when only using 1\% of EuroSat2k, our multi-modal pretext tasks improve LP performance from 16.5\% to 34\% accuracy. 

\begin{table}[tbp]
\centering
\caption{\textbf{Multi-modal pretext tasks.}
Fine-tuning (FT) and linear probing (LP) performance
for multi-pretext pretraining (MMEarth64, MMEarth) and masked image pretraining (ImageNet, MMEarth64-S2). We also present multi-pretext results with subsets of modalities using only pixel-level (MMEarth64-PixelM) or only image-level (MMEarth64-ImageM) modalities in addition to the Sentinel-2 optical image reconstruction. Multi-pretext pretraining improves both FT and LP results. }
\label{tab:results_multi_modal}
\adjustbox{max width=\textwidth}{
\begin{tabular}{lcccc}
\toprule
Pretrain       & \begin{tabular}[c]{@{}c@{}}BigEarth20k\\(F1$\uparrow$)\\ FT/LP\end{tabular} & \begin{tabular}[c]{@{}c@{}}So2Sat20k\\(Acc.$\uparrow$)\\ FT/LP\end{tabular} & \begin{tabular}[c]{@{}c@{}}Cashew1k\\(IoU$\uparrow$)\\ FT\end{tabular} & \begin{tabular}[c]{@{}c@{}}SAcrop3k\\(IoU$\uparrow$)\\ FT\end{tabular} \\ \midrule
ImageNet              & 55.7/25.9 & 36.6/24.0 &      77.1 &      26.7 \\
MMEarth64-S2          & 65.1/36.2  & 48.7/33.5     & 79.9  &     36.0  \\ \midrule
MMEarth64-PixelM      & 67.5/36.6 & \textbf{58.2}/38.5 &      \textbf{81.9} &      39.7  \\
MMEarth64-ImageM      & \textbf{68.4}/36.6 & 56.3/39.2 &      81.4 &      \textbf{39.8}  \\
MMEarth64             & 68.2/40.9 & 54.0/41.7 &      81.6 &      39.7  \\
MMEarth               & 67.1/\textbf{43.3}  & 54.6/\textbf{43.8}  & 79.8  & 38.2  \\ \bottomrule

\end{tabular}
}
\end{table}

\noindent\textbf{Compares favorably to prior work.} 
We compare our MP-MAE method to a small, but most closely related, subset of prior work in Table~\ref{tab:results_prior_work}. While all models rely on different pretraining datasets and pretraining strategies, we show that our model compares favorably to prior work on most tasks.
The encoder in our MP-MAE consists of 3.7M parameters, which is much smaller compared to e.g. the 25M for ResNet-50.
This supports our motivation for creating better domain-specific pretraining datasets and pretraining strategies.

\begin{table}[tbp]
\centering
\caption{\textbf{Comparison to prior work.} We compare our multi-pretext pretraining with prior work that makes use of different pretraining datasets and strategies. Our pretraining strategy achieves similar if not better performance compared to other methods that use a ResNet-50 backbone. We highlight the \textbf{best} and \underline{second best} results.}
\label{tab:results_prior_work}
\adjustbox{max width=\textwidth}{
\begin{tabular}{@{}lllcccc@{}}
\toprule
Method         & \begin{tabular}[l]{@{}c@{}}Pretrain data\\(Num. locations) \end{tabular}            & \begin{tabular}[l]{@{}c@{}}Encoder\\(Num. parameters)  \end{tabular}      & \begin{tabular}[l]{@{}c@{}}BigEarth20k\\(F1$\uparrow$) \\ FT/LP \end{tabular} & \begin{tabular}[c]{@{}c@{}}So2Sat20k\\(Acc.$\uparrow$)\\ FT/LP \end{tabular} & \begin{tabular}[c]{@{}c@{}}Cashew1k\\(IoU$\uparrow$)\\ FT\end{tabular} & \begin{tabular}[c]{@{}c@{}}SAcrop3k\\(IoU$\uparrow$)\\ FT\end{tabular} \\ \midrule
GASSL~\cite{ayush2021geography}          & fMoW (0.36M)          & ResNet-50 (25M) & 53.6/34.1 & 42.4/35.4 &      \underline{79.0} &      21.1  \\
SeCo~\cite{manas2021seasonal}           & SeCo (0.20M)          & ResNet-50 (25M)       & \underline{59.6}/\textbf{44.5} & \underline{45.2}/\underline{39.9} &      77.5 &      \underline{22.7}  \\
SatlasNet~\cite{bastani2023satlaspretrain} & SatlasPretrain (0.86M) & ResNet-50 (25M) & 55.5/40.3 & 37.1/22.1 &      76.4 &      20.5  \\
MP-MAE (ours)  & MMEarth (1.24M)        & ConvNeXt V2-Atto (3.7M) & \textbf{67.1}/\underline{43.3}  & \textbf{54.6}/\textbf{43.8}  & \textbf{79.8}  & \textbf{38.2}  \\
\bottomrule
\end{tabular}
}
\end{table}

\subsection{Limitations}
While this study introduces a new dataset and methodology to advance the use of multiple modalities for representation learning, some limitations exist.
First, we rely on GEO-Bench for downstream evaluation with five datasets covering a range of tasks and geographic regions. Further evaluation on diverse application domains from various geographical regions could reveal insights about which scenarios our proposed multi-modal pretext tasks are beneficial in at a global scale.
Second, we run our pretraining for 200 epochs due to restricted compute resources. For the same reason we only train encoders with a limited number of parameters. Prior work exploring MAEs for representation learning pretrains for 800 or 1600 epochs. 
Third, our MP-MAE is not explicitly trained to be robust to varying numbers of input channels. We have trained the MP-MAE with multi-spectral Sentinel-2 images (12 bands). It needs to be investigated how our models transfer to downstream tasks that use e.g. RGB only. While this seems to be an artefact of benchmark datasets, rather than a real-world scenario when working with Sentinel-2 images, prior work proposed input masking strategies to make a model robust against missing channels~\cite{tseng2023lightweight}, which could be built into our framework.
Fourth, we are not making use of the validation dataset for early stopping during the FT/LP evaluation, which could improve results overall. 
Lastly, we only reported single trials. Running the FT/LP experiments multiple times would allow us to average out random fluctuations.

\section{Conclusion}
Prior work has mostly focused on exploiting \emph{geolocation} and \emph{time} for representation learning with EO data. We move a step forward and present MMEarth, a global \emph{multi-modal} pretraining dataset for geospatial representation learning. With 12 modalities sampled from 1.2M locations, we propose a Multi-Pretext Masked Autoencoder (MP-MAE) approach based on ConvNeXt V2 to study the potential of aligned multi-modal data to learn better representations for Sentinel-2 optical satellite images.
First, we show how to adapt the ConvNeXt V2 encoder to improve domain-specific pretraining with medium-resolution satellite images.
Second, we demonstrate that multi-modal pretext tasks are beneficial for learning better representations for Sentinel-2. Both fine-tuning and linear probing are improved; the latter profits substantially in the few-shot setting. This is promising for real-world applications with limited training data.
However, further research is needed to close the gap between the fine-tuning and linear probing performance. 

\subsubsection{Acknowledgments.} 
We thank Lucia Gordon for the valuable feedback.
We greatly appreciate the open data policies of the Copernicus program and its partners ESA and ECMWF. We thank Google Earth Engine for hosting the data and providing free access.
This work was supported in part by the Pioneer Centre for AI, DNRF grant number P1.
The authors AK, CI, and NL acknowledge support by the research grant DeReEco (grant number 34306) from Villum Foundation.
SO and CI acknowledge support by the research grant Global Wetland Center (grant number NNF23OC0081089) from Novo Nordisk Foundation.
CI and SB acknowledge  support by  the European Union project ELIAS (grant agreement number 101120237). 
We thank the Danish e-Infrastructure Consortium (DeiC), Martin Brandt, and Konrad Schindler for their support with computing resources.


%
%
\bibliographystyle{splncs04}
\bibliography{main}

\clearpage
\appendix
\setcounter{table}{0}
\renewcommand{\thetable}{A\arabic{table}}
\setcounter{figure}{0}
\renewcommand{\thefigure}{A\arabic{figure}}

\section{Dataset details for MMEarth}
MMEarth was collected at 1.2M locations around the world. The data was first downloaded at tile size of 1300m$\times$1300m corresponding to approximately $130\times130$ pixels. As this can lead to slight variations by $\pm$1 pixel, we center cropped to $128\times128$ pixels to harmonize the final tile size. The total dataset is 639GB in size and offered as a compressed gzip hdf5 file (lossless compression).

\begin{figure}[]
  \centering
  \begin{subfigure}{0.48\linewidth}
    \includegraphics[width=\linewidth]{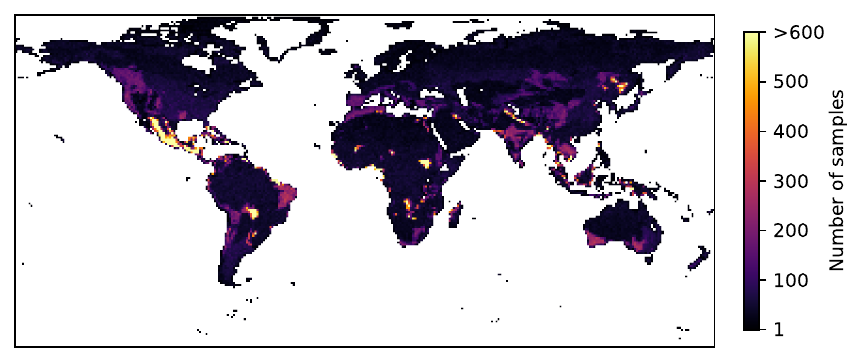}
    \caption{Spatial distribution L1C}
    \label{fig:distribution_spatial_l1c}
  \end{subfigure}
  \hfill
  \begin{subfigure}{0.48\linewidth}
    \includegraphics[width=\linewidth]{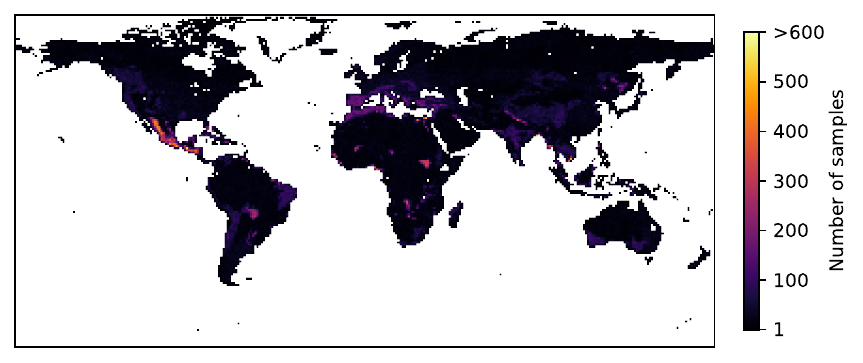}
    \caption{Spatial distribution L2A}
    \label{fig:distribution_spatial_l2a}
  \end{subfigure}
  \caption{Spatial distribution of L1C and L2A data.}
  \label{fig:data_distribution1}
  
\vspace{1em}

  \begin{subfigure}[b]{0.48\linewidth}
    \includegraphics[width=\linewidth]{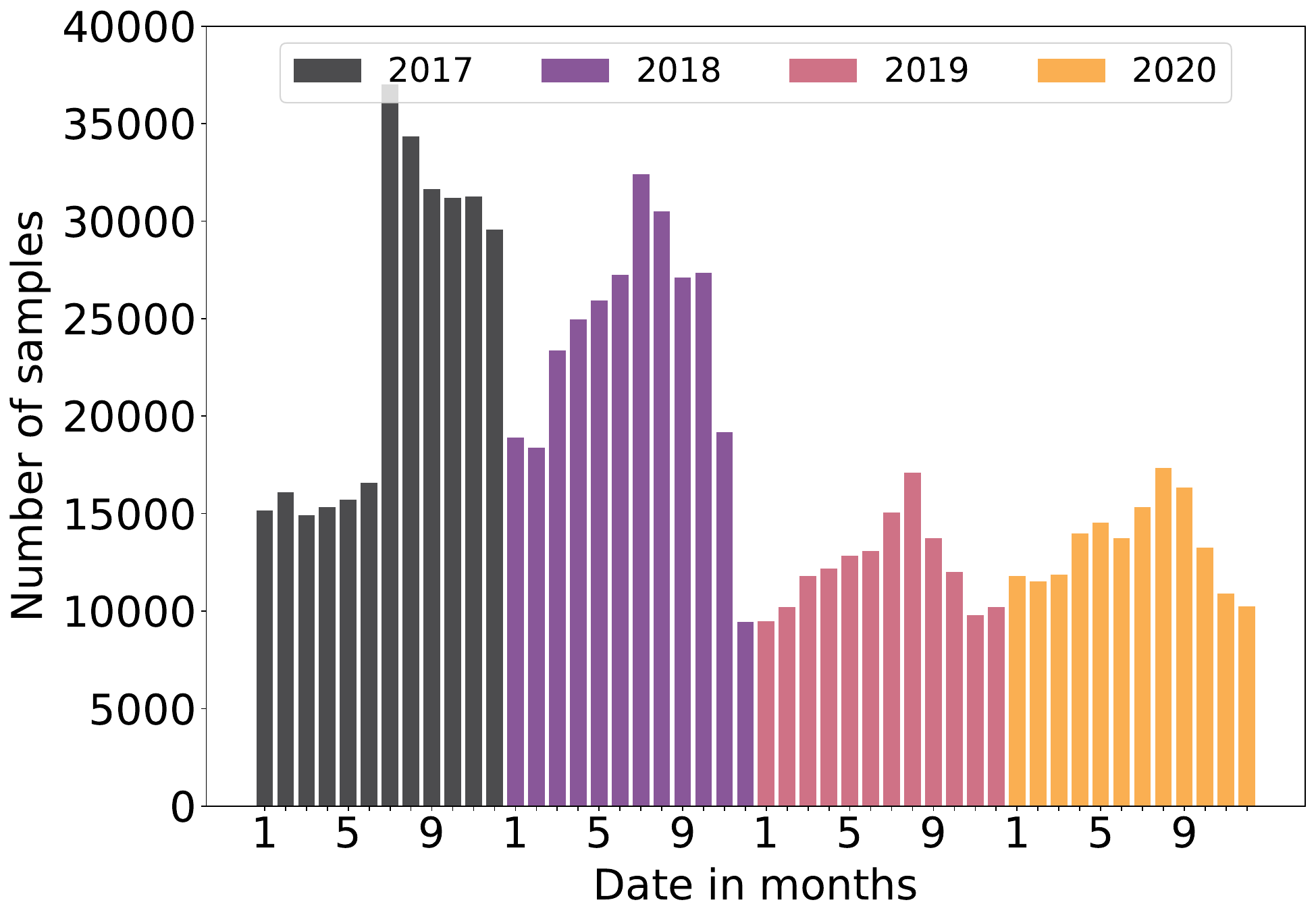}
    \caption{Temporal distribution L1C}
    \label{fig:distribution_temporal_l1c}
  \end{subfigure}
  \hfill
  \begin{subfigure}[b]{0.48\linewidth}
    \includegraphics[width=\linewidth]{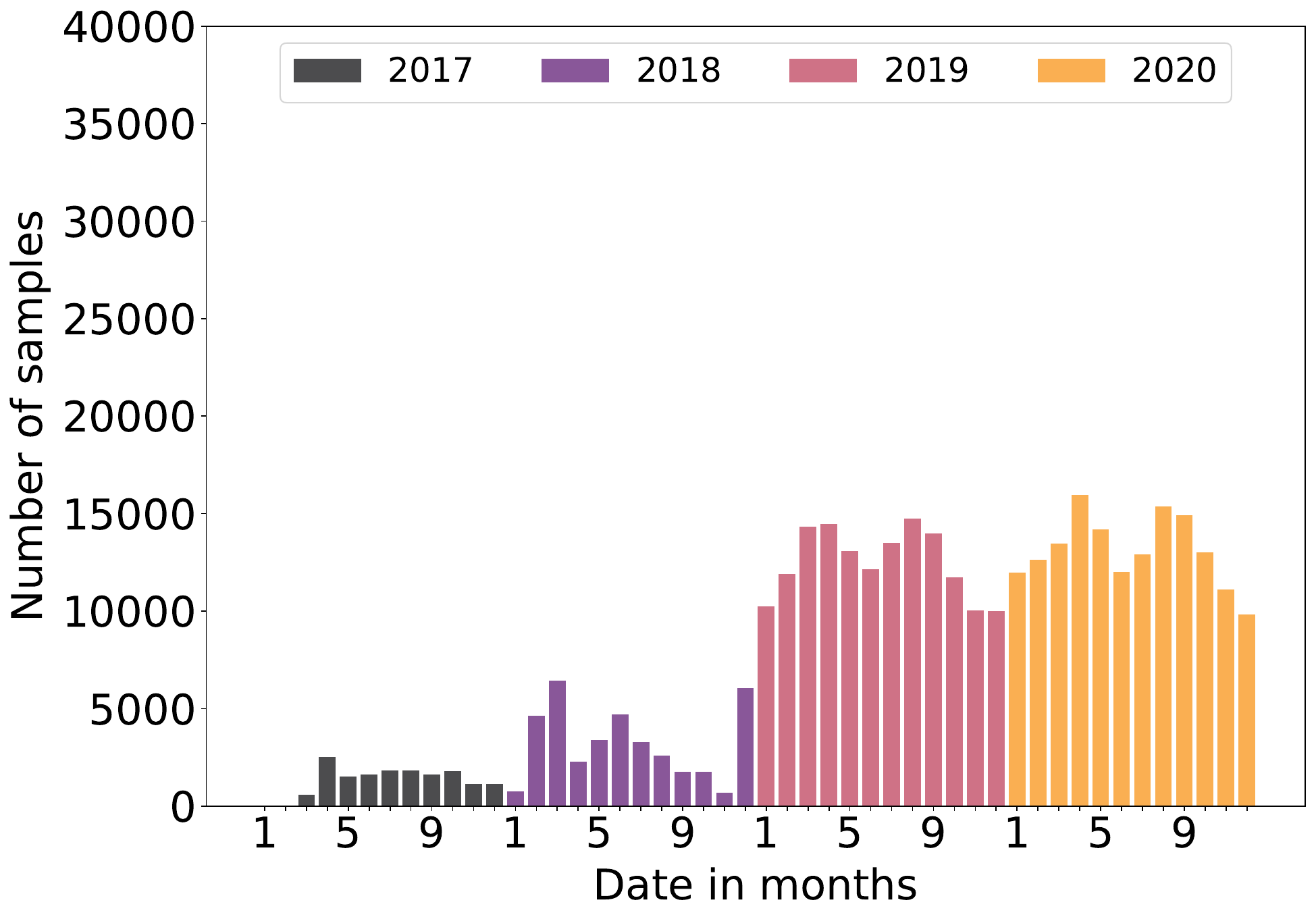}
    \caption{Temporal distribution L2A}
    \label{fig:distribution_temporal_l2a}
  \end{subfigure}
  \caption{Temporal distribution of l1c and l2a data.}
  \label{fig:data_distribution2}
\end{figure}

Fig.~\ref{fig:data_distribution1} and Fig.~\ref{fig:data_distribution2} show the spatial and temporal distribution of Sentinel-2 L1C (top of atmosphere) and L2A (bottom of atmosphere) data. The atmospherically corrected L2A data is globally available on Google Earth Engine \cite{gorelick2017google} (GEE) starting November 2018 until today. Before November 2018, L2A only exists in certain parts of the world (\eg Europe).  Hence, to get a uniform distribution of samples throughout the four years 2017-2020, we ensured that L1C is sampled more frequently between January 2017 and November 2018. From December 2018, we have randomly sampled L1C or L2A data to guarantee that  later years are also covered by both product levels.

\begin{figure}[tb]
  \centering
  \begin{subfigure}{0.45\linewidth}
    \includegraphics[width=\linewidth]{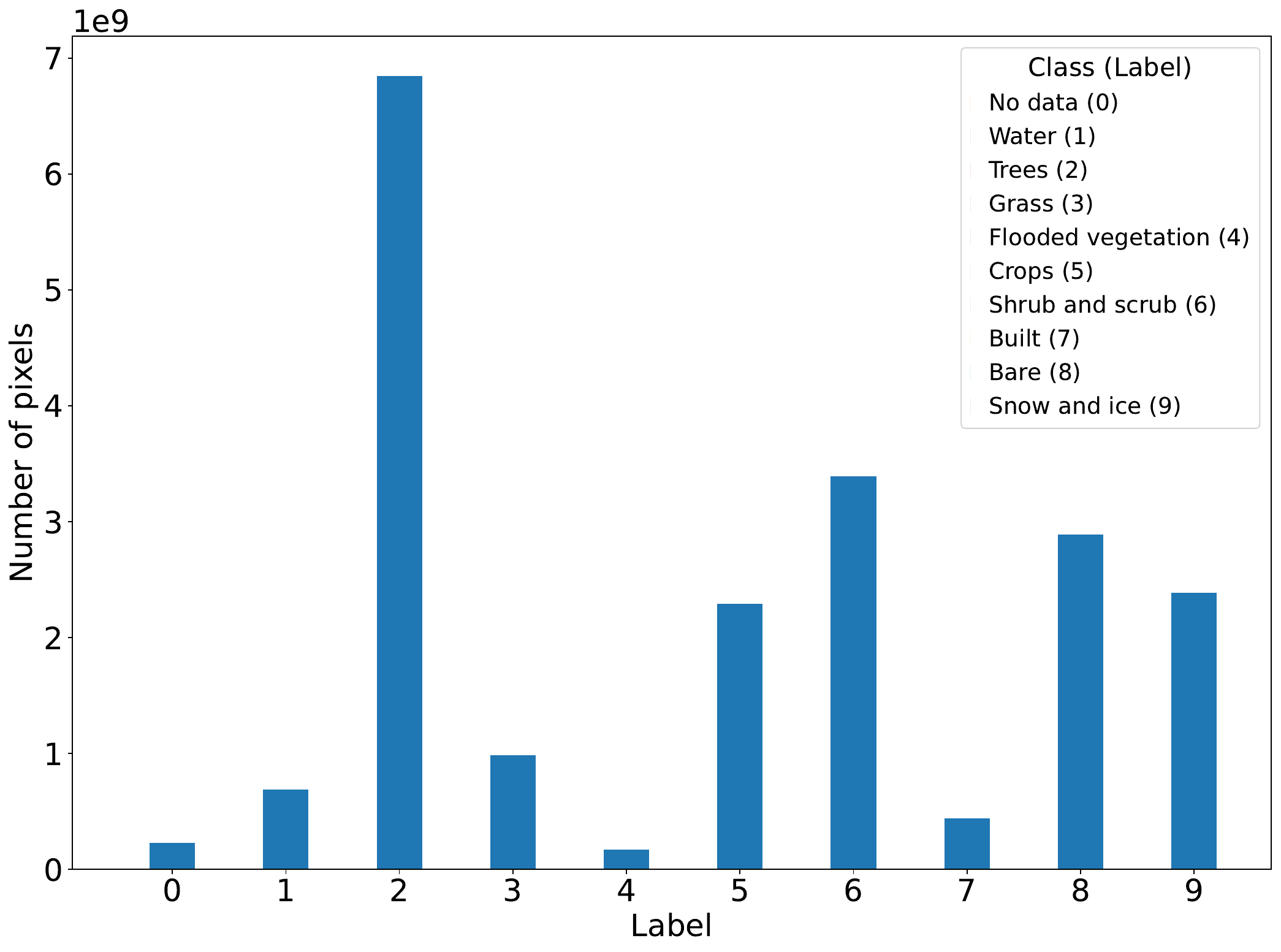}
    \caption{Dynamic World landcover}
    \label{fig:dynamic_world}
  \end{subfigure}
  \hfill
  \begin{subfigure}{0.45\linewidth}
    \includegraphics[width=\linewidth]{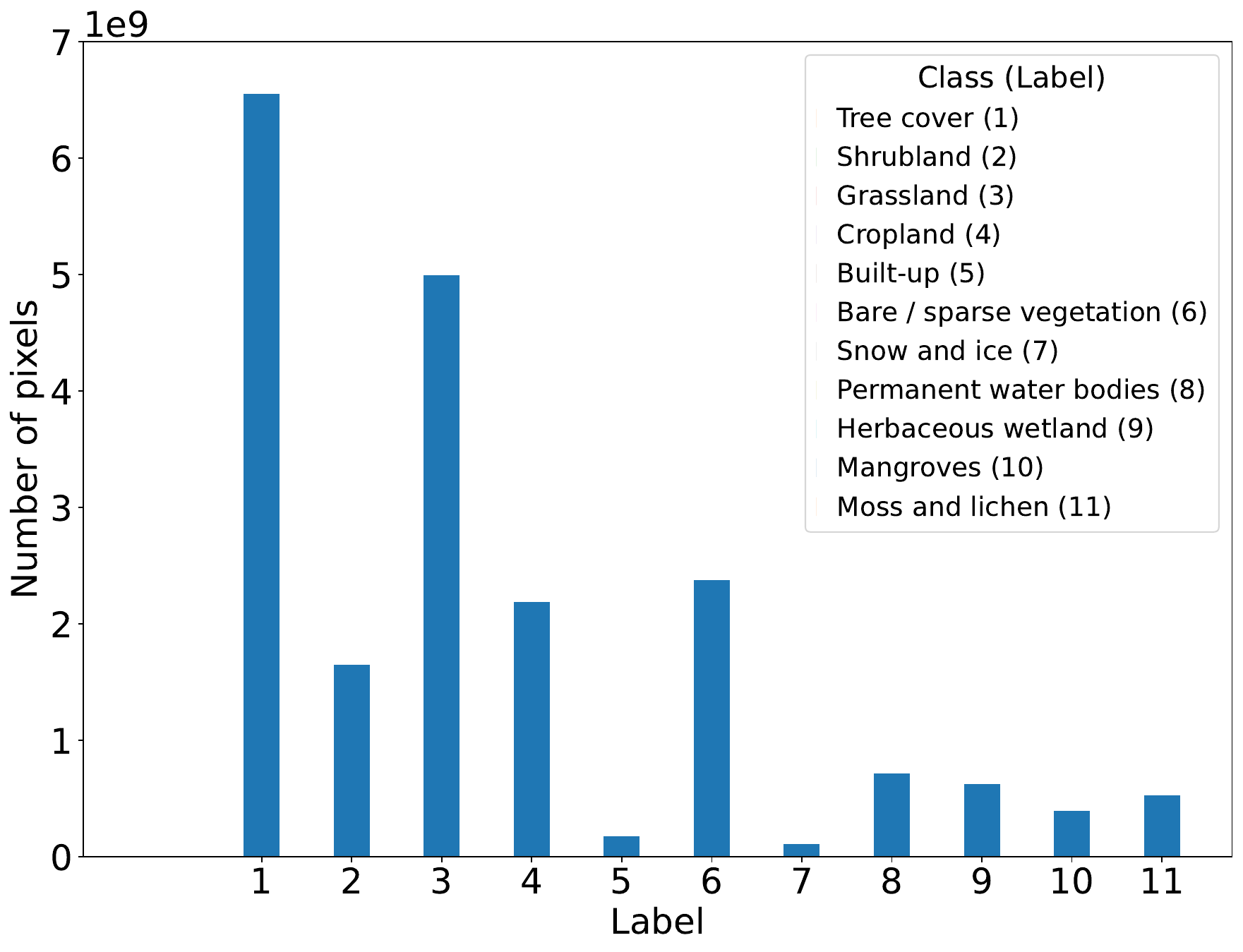}
    \caption{ESA World Cover}
    \label{fig:esa}
  \end{subfigure}
  \hfill
  \begin{subfigure}{0.45\linewidth}
    \includegraphics[width=\linewidth]{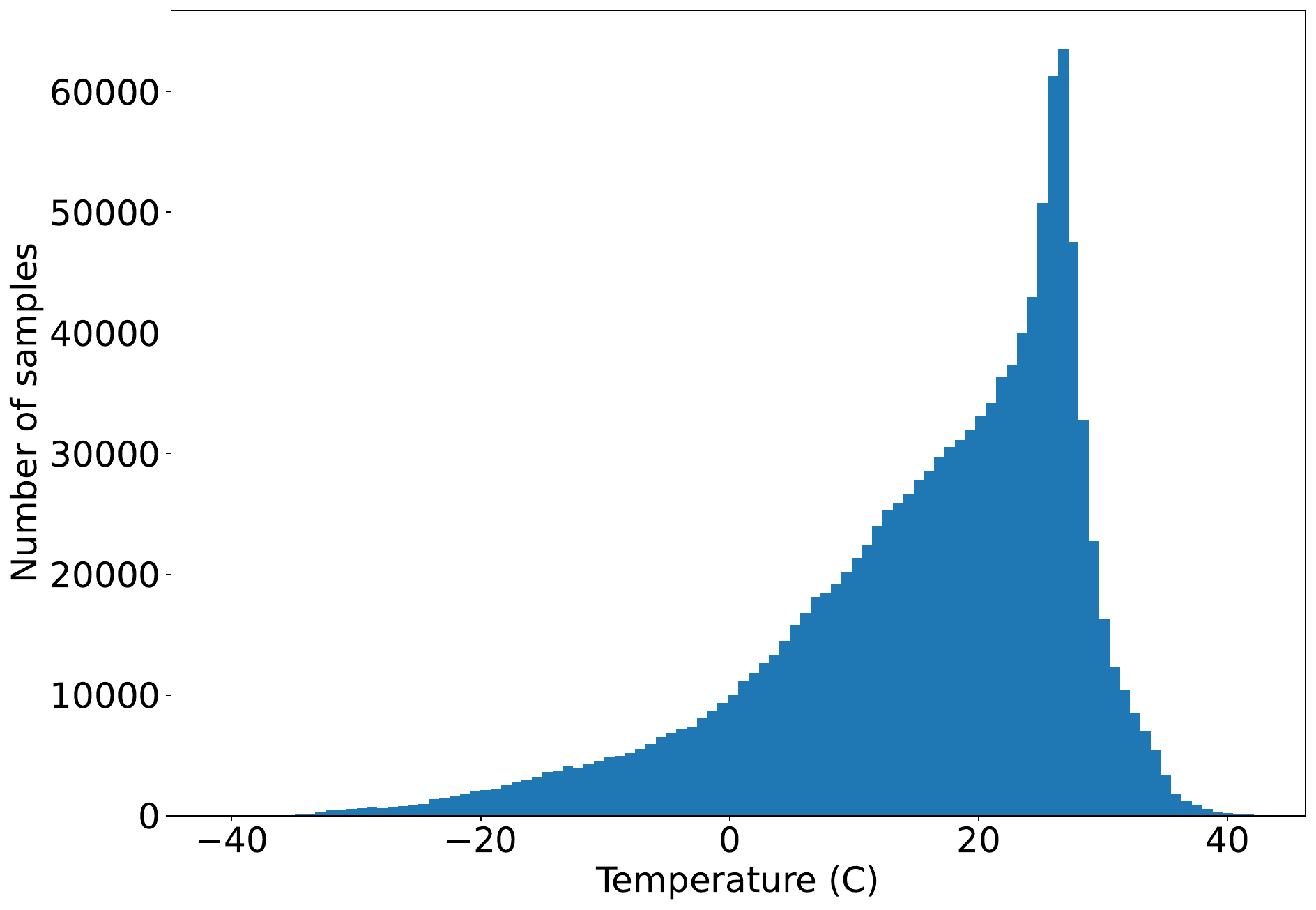}
    \caption{ERA5 mean temp.~of current month}
    \label{fig:era5}
  \end{subfigure}
  \hfill
  \begin{subfigure}{0.45\linewidth}
    \includegraphics[width=\linewidth]{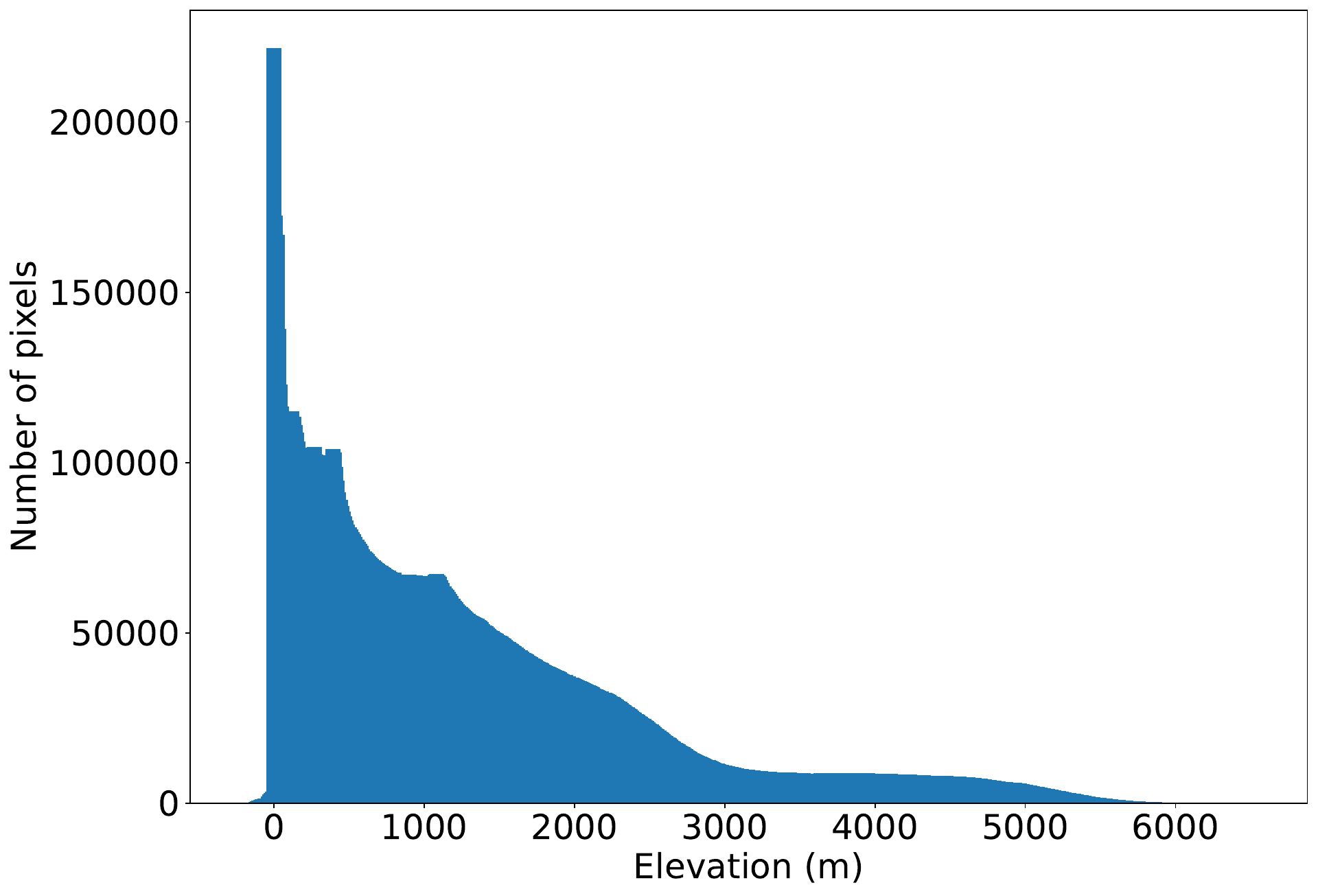}
    \caption{Aster elevation}
    \label{fig:aster}
  \end{subfigure}
  
  \caption{\textbf{Distribution of additional modalities.}
    }
    
  \label{fig:target_distribution}
\end{figure}

The distribution of additional modalities is visualized in
Fig.~\ref{fig:target_distribution}. Due to our biome-balanced sampling scheme, described in the following section, the MMEarth dataset focuses on the natural world, but also contains some built-up landcover. Thus, MMEarth was designed as a pretraining dataset for applications in environmental monitoring.

\subsection{Sampling strategy}

\subsubsection{Sampling of locations.}

MMEarth was collected, ensuring uniform sampling across 14 biomes \cite{Dinerstein2017}. Each biome can further be divided into a number of ecoregions. 
We used the GEE toolbox for this stratified sampling approach. However, since the area of the largest biome was too large to be processed by GEE, we facilitate stratified sampling by considering the area within an ecoregion. The number of samples per ecoregion was computed by
\begin{equation}
    N_e=\frac{N_t}{14} \cdot \frac{A_e}{A_{B_{e}}} ~,
\label{eq:uniform_biomes}
\end{equation}
where $N_e$ is the number of samples for an ecoregion, $N_t$ is the total number of samples to collect (1.2M in our case), $A_e$ is the area covered by that ecoregion, $A_{B_{e}}$ is the area covered by the biome the ecoregion is part of and $14$ represents the number of biomes. This process was repeated for all ecoregions and for all biomes to finally obtain a uniform sample across the biomes. 

\subsubsection{Sampling of Sentinel-2.} For every location, we first randomly chose a year from 2017-2020. For the selected year, we randomly sampled either L1C and L2A. If no L2A data existed, we sampled L1C data (since L2A is not globally available until  November 2018). 
We queried the respective Sentinel-2 data collection with \emph{CLOUD\_PIXEL\_PERCENTAGE} of less than 10\% at the original 100km$\times$100km tile level. 
We discarded images that contain no data values at the boundary of the orbit coverage.
From this collection of candidates, we randomly sampled one Sentinel-2 image date. 
Finally we cropped the respective tile centered at the query location.
We downloaded all bands (\ie 13 for L1C and 12 for L2A) and reprojected all bands to a 10m resolution using bilinear interpolation. For our experiments, we always used the 12 bands corresponding to the L2A product, excluding band B10 in L1C.

We found that the cloud filter at the original tile level mostly led to cloud free samples, but some samples contained clouds. 
In our experiments, we did not apply any cloud filtering at the sample level, although that would be possible with the information provided in MMEarth.

\subsubsection{Sampling and processing of additional modalities.} In this sub-section, we explain any additional processing done when sampling modalities. We only focus on those modalities that require additional processing apart from the ones performed when requesting for data using the GEE API. All pixel-level modalities were reprojected to 10m Sentinel-2 grid using bilinear interpolation.

\textbf{Sentinel-1.} Sentinel-1 data was sampled with reference to the sampled Sentinel-2 observation date. Hence, we sampled the closest Sentinel-1 image from both the ascending and descending orbits. We downloaded all four available bands. 

\textbf{Aster GDEM.} Aster data from GEE includes an elevation band, from which we have computed the slope.

\textbf{Dynamic World.} We first sampled the dynamic world collection for the full year of the Sentinel-2 observation date. The data from GEE comes with labels from 0 to 8. We also realized that \emph{NO\_DATA} values are also mapped to 0. Hence, we mapped the labels from 0 to 9, and indicated 0 to be the \emph{NO\_DATA} pixels. We then compute the mode of the full collection. This was done since dynamic world images for a specific date can sometimes lead to \emph{NO\_DATA} values, and hence to reduce the amount of such pixels, we computed the mode of a set of images. 
 
\textbf{ERA5.} We selected 4 bands from the ERA5 dataset on GEE (average temperature, minimum temperature, maximum temperature and total precipitation). These statistics are monthly. Additionally, we computed the yearly average, minimum, maximum, and total, respectively, for each of these statistics. 
We collected 3 sets of statistics for the same month, previous month, and full year corresponding to the Sentinel-2 observation date ($3 \times 4\ \text{bands} = 12\ \text{bands}$). 

\newcommand{\Lat}{\text{Lat}}
\renewcommand{\Lat}{\phi}
\newcommand{\Lon}{\text{Lon}}
\renewcommand{\Lon}{\lambda}

\textbf{Geolocation.} We use a cyclic encoding of the latitude $\Lat$ and longitude $\Lon$ corresponding to the center point as follows:
\begin{equation}
    \Lat_{\sin} = \sin{(2\pi \Lat / 360)}
\end{equation}
\begin{equation}
    \Lat_{\cos} = \cos{(2\pi \Lat / 360)}
\end{equation}
    
\begin{equation}
    \Lon_{\sin} = \sin{(2\pi \Lon / 360)}
\end{equation}
\begin{equation}
    \Lon_{\cos} = \cos{(2\pi \Lon / 360)}
\end{equation}

\textbf{Date.} Similarly, we use a cyclic encoding of the month of the Sentinel-2 observation date:

\newcommand{\Month}{\text{month}}
\renewcommand{\Month}{m}

\begin{equation}
    \Month_{\sin} = \sin{(2\pi m / 12)}
\end{equation}
\begin{equation}
    \Month_{\cos} = \cos{(2\pi m / 12)}
\end{equation}
    Here $m$ corresponds to the month as an integer from 1 to 12.

\subsection{Licenses of data sources}
MMEarth is constructed from publicly available data distributed under non-restrictive licenses. We list the license for every data source in Table~\ref{tab:licenses}. The MMEarth dataset is released under the CC BY 4.0 license.

\begin{table}[]
\caption{\textbf{Licenses of data sources.} All collected data is licensed for open usage. The detailed license information is listed for each data source. All urls were accessed 2024-03-11.}
\adjustbox{max width=\textwidth}{
\begin{tabular}{@{}lll@{}}
\toprule
Data source     & License                                                         & Info                                                                           \\ \midrule
Sentinel-2      & CC BY-SA 3.0 IGO                               & \url{https://open.esa.int/copernicus-sentinel-satellite-imagery-under-open-licence/} \\
Sentinel-1      & CC BY-SA 3.0 IGO                               & \url{https://open.esa.int/copernicus-sentinel-satellite-imagery-under-open-licence/} \\
Aster GDEM v3   & similar to CC0                                 & \url{https://lpdaac.usgs.gov/data/data-citation-and-policies/}                       \\
ETH-GCHM        & Creative Commons Attribution 4.0 International & \url{https://langnico.github.io/globalcanopyheight}                                 \\
Dynamic World   & Creative Commons BY-4.0                        & \url{https://dynamicworld.app/about/}                                                \\
ESA World Cover & Creative Commons Attribution 4.0 International & \url{https://esa-worldcover.org/en/data-access\#citation}                            \\
Biome           & Creative Commons Attribution 4.0 International & \url{https://ecoregions.appspot.com/}     \\
Ecoregion       & Creative Commons Attribution 4.0 International & \url{https://ecoregions.appspot.com/}                                                \\
ERA5            & Copernicus C3S/CAMS License agreement          & \url{https://www.ecmwf.int/en/forecasts/dataset/ecmwf-reanalysis-v5}                 \\ \bottomrule
\end{tabular}
}
\label{tab:licenses}
\end{table}

\section{Implementation details}
\noindent\textbf{Pretraining:} We mainly followed the ConvNeXt V2 hyperparameter settings. If not stated otherwise, we pretrained for 200 epochs with a base learning rate 1.5$\cdot10^{-4}$ and an effective batch size of $4096$. The only data augmentation used was random cropping. 
We standardize each channel in the input and targets to zero mean and unit variance on the pretraining data. 
To harmonize the two Sentinel-2 products, we standardize L1C and L2A individually.
The target normalization harmonizes the losses for modalities with shared characteristics (\eg continuous targets). Missing input pixels are replaced by zeros (\ie the mean). 
Local target patch normalization is used for reconstructing the optical image bands~\cite{He_2022_CVPR}. 

\noindent\textbf{Fine-tuning parameters:}
We fine-tuned and linear probed for 100 epochs, and maintained an effective batch size of 1024 with a base learning rate of 2$\cdot 10^{-4}$. 
For the semantic segmentation tasks we fine-tuned (FT) in two phases. First, we trained the randomly initialized U-Net decoder with the frozen pretrained encoder for 50 epochs, and then fine-tuned the full model for another 150 epochs. We maintained an effective batch size of 32, and a base learning rate of 0.01. We did not apply any data augmentation.
For all FT/LP experiments we report results on the test split using the checkpoint of the last epoch.

\noindent\textbf{Semantic segmentation:}
We adopt a U-Net~\cite{ronneberger2015u} architecture to evaluate the pretrained encoders. 
The upsampling block consists of a nearest neighbour upsampling followed by a convolutional layer with 3$\times$3 filter kernels, layer norm and a GELU activation layer. 

\noindent\textbf{Input bands for So2Sat20k:}
So2Sat20k only provides 10 instead of 12 bands. However, to make use of these additional multi-spectral bands, we filled in the missing bands (B1 and B9) with a copy of the band with the closest wavelength. I.e., we use a copy of B2 for B1, and B8A for B9.
This could lead to a lower linear probing performance but should not affect results much when fine-tuning the entire encoder.

\section{Taster pretraining datasets: MMEarth100k and MMEarth64}

To facilitate research in self-supervised learning with limited compute resources, we provide two subsets of the MMEarth dataset: 
MMEarth100k and MMEarth64.
The MMEarth100k is a random subset of 100k locations with the full 128$\times$128 pixels and has a data volume of roughly 48GB.
The MMEarth64 contains all 1.2M locations, but all modalities are center cropped to obtain rasters of 64$\times$64 pixels, which reduces the data volume roughly by factor 4 from 597GB to 152GB.  

We provide results using these taster subsets to pretrain our MP-MAE approach with the same setting as used for pretraining on the full MMEarth dataset in Table~\ref{tab:taster_results}. 
The only change required for MMEarth64 is the patch size. As the image size is halved from 112 to 56 pixels, we also half the patch size from 16 to 8 pixels which keeps the number of patches the same. 
For the downstream tasks that we consider in this evaluation, we observe that pretraining on MMEarth64 performs comparably to MMEarth, and even slightly better in some tasks. However, there might be downstream tasks that could profit from larger pretraining image sizes. Also alternative self-supervised pretraining strategies might benefit from having access to the larger input sizes.
Pretraining on MMEarth100k leads to a consistent drop in performance, but still outperforms the ImageNet pretrained models on all tasks except the Cashew1k.
From these initial results, we conclude that the diversity and number of samples are more important pretraining characteristics than the size of the input images for learning generalizing representations with our proposed MP-MAE.
\begin{table}[tb]
\centering
\caption{\textbf{MP-MAE results on MMEarth taster datasets.}}
\label{tab:taster_results}
\adjustbox{max width=\textwidth}{
\begin{tabular}{lllcccc}
\toprule
Pretrain data & \begin{tabular}[l]{@{}c@{}}Image \\size\end{tabular} & \begin{tabular}[l]{@{}c@{}}Patch \\size\end{tabular} & \begin{tabular}[l]{@{}c@{}}BigEarth20k\\(F1$\uparrow$) \\ FT/LP \end{tabular} & \begin{tabular}[c]{@{}c@{}}So2Sat20k\\(Acc.$\uparrow$)\\ FT/LP \end{tabular} & \begin{tabular}[c]{@{}c@{}}Cashew1k\\(IoU$\uparrow$)\\ FT\end{tabular} & \begin{tabular}[c]{@{}c@{}}SAcrop3k\\(IoU$\uparrow$)\\ FT\end{tabular} \\ \midrule
ImageNet      & 224   & 32    & 55.7/25.9 & 36.6/24.0 &      77.1 &      26.7 \\
MMEarth       & 112   & 16    & 67.1/\textbf{43.3}  & \textbf{54.6}/\textbf{43.8}  & 79.8  & 38.2  \\
MMEarth100k   & 112   & 16    & 59.8/35.4 & 38.8/31.2 &      67.5 &      32.9  \\
MMEarth64     & 56    & 8     & \textbf{68.2}/40.9 & 54.0/41.7 &      \textbf{81.6} &      \textbf{39.7}  \\
\bottomrule
\end{tabular}
}
\end{table}

\section{Additional results}

\subsection{Ablation of multi-pretext loss strategy}
Results from Table~\ref{tab:loss_weighting_results} show that for the Atto model, pretraining with uncertainty based loss weighting achieves comparable performance to equal weighting. While the performance is comparable, we note that uncertainty loss weighting provides additional insights on the importance of tasks during pretraining (see Fig.~\ref{fig:uncertainty_plots}). 

\subsubsection{Analysis of the estimated task uncertainty}

\begin{figure}[tb]
  \centering
  \begin{subfigure}{0.45\linewidth}
    \includegraphics[width=\linewidth]{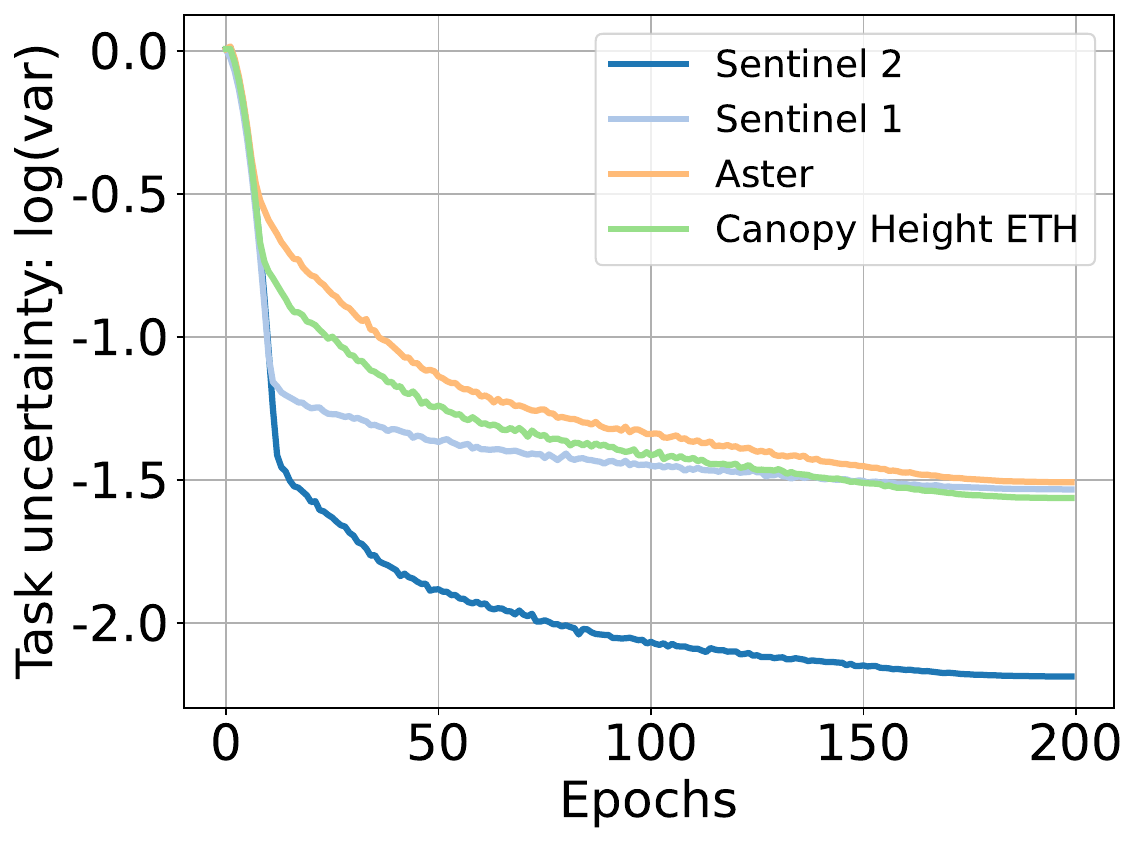}
    \caption{Pixel-level continuous modalities}
  \end{subfigure}
  \hfill
  \begin{subfigure}{0.45\linewidth}
    \includegraphics[width=\linewidth]{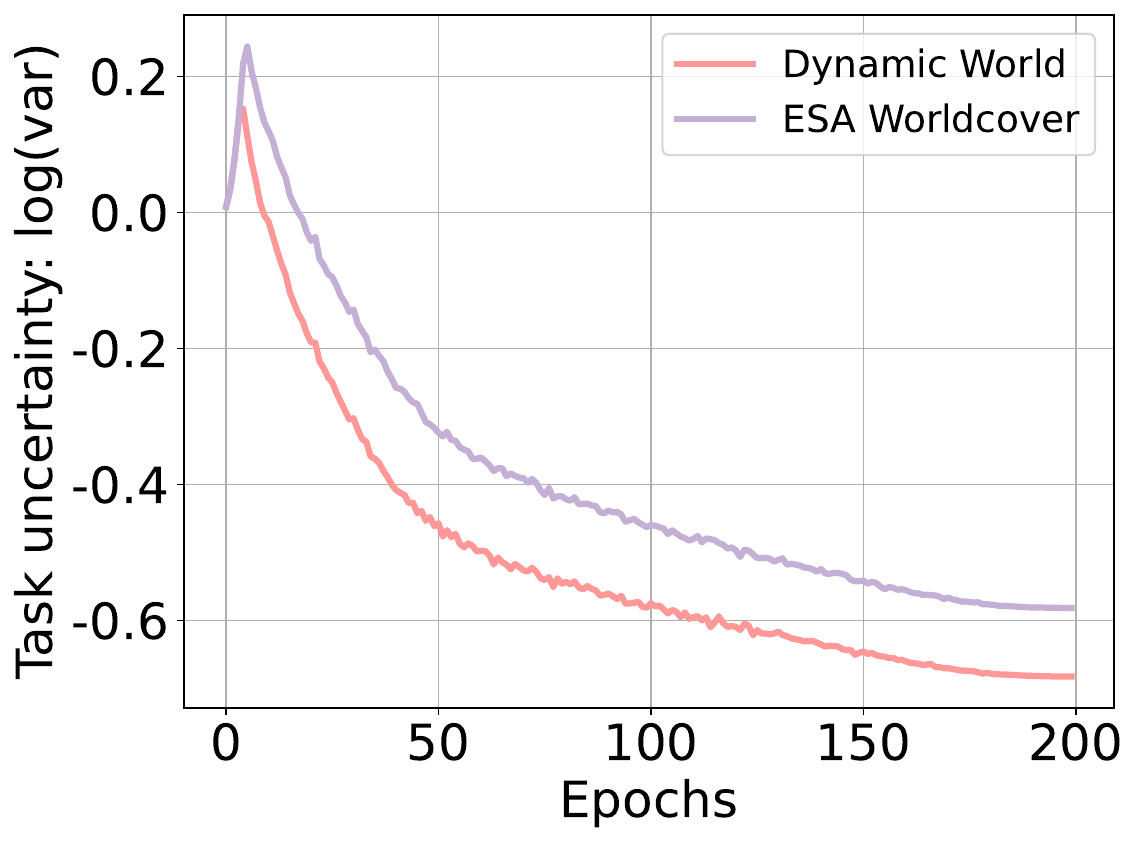}
    \caption{Pixel-level categorical modalities}
  \end{subfigure}
  \hfill
  \begin{subfigure}{0.45\linewidth}
    \includegraphics[width=\linewidth]{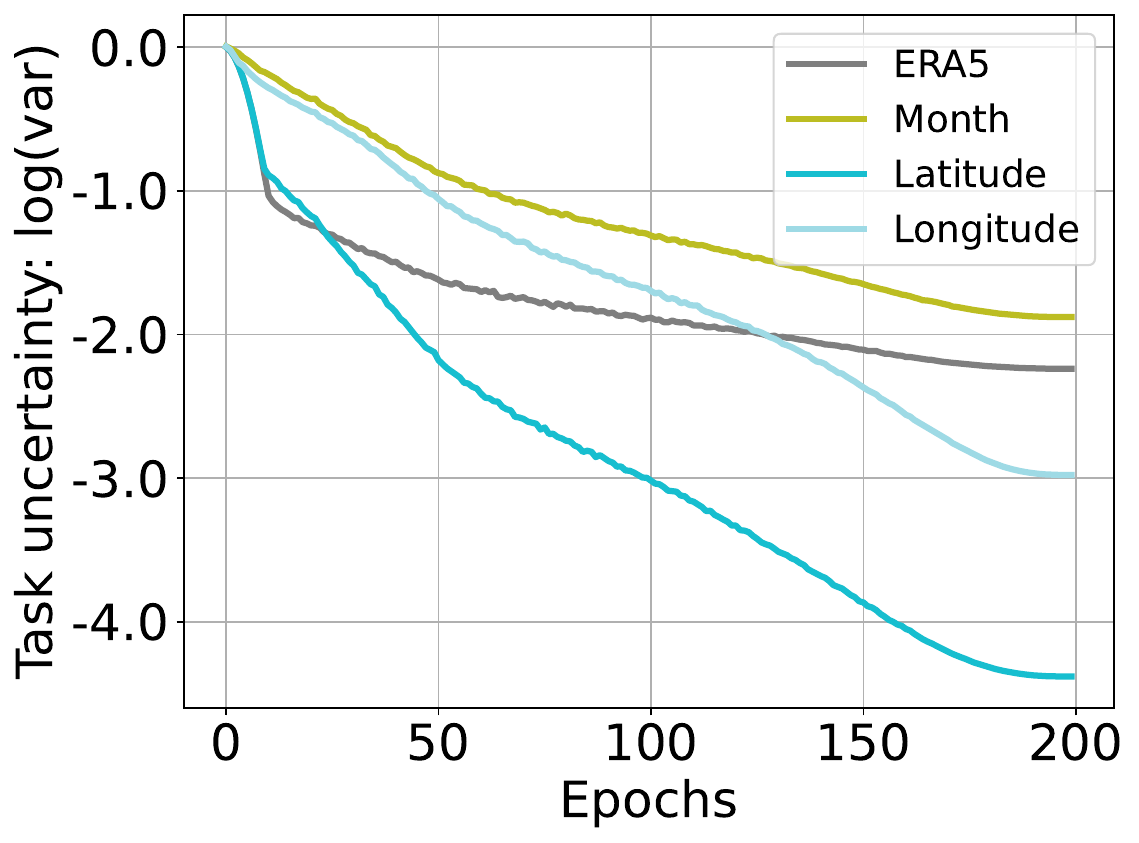}
    \caption{Image-level continuous modalities}
  \end{subfigure}
  \hfill
  \begin{subfigure}{0.45\linewidth}
    \includegraphics[width=\linewidth]{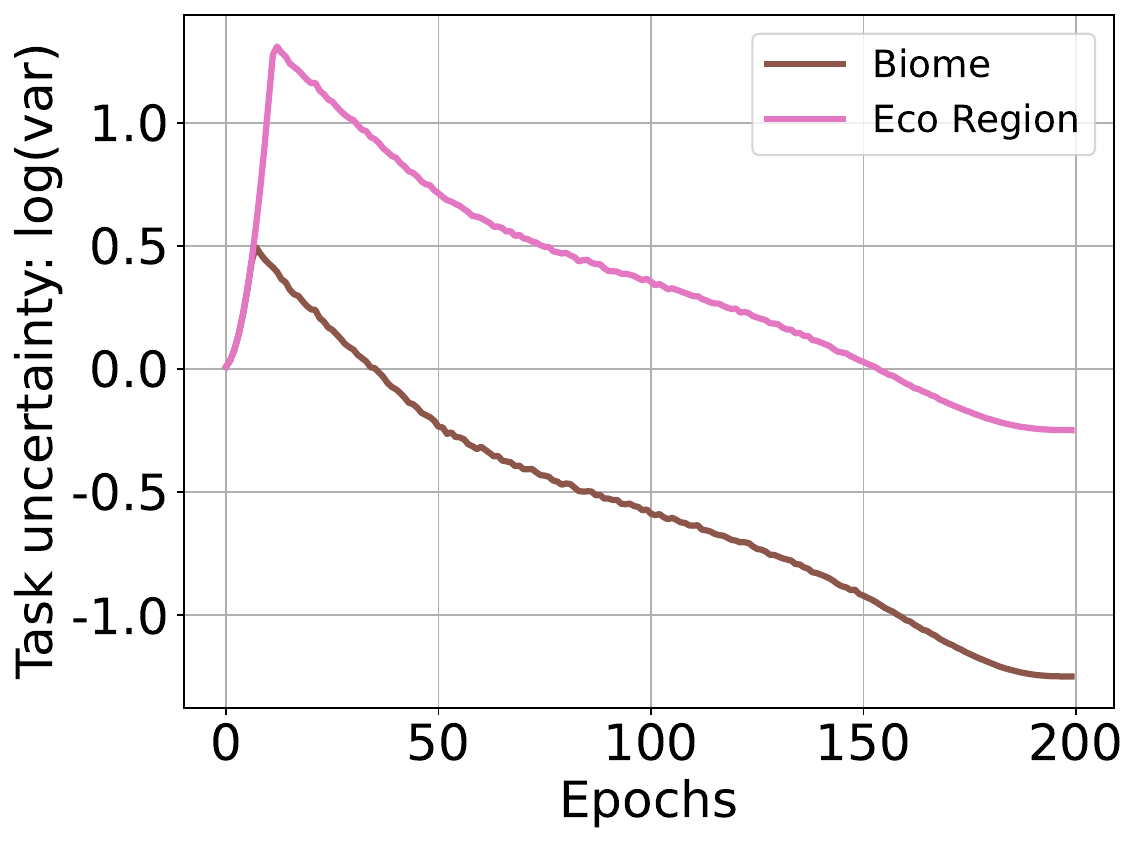}
    \caption{Image-level categorical modalities}
  \end{subfigure}
  
  \caption{
    \textbf{Task uncertainty during pretraining.} We plot the $log(var)$ for each task grouped by pixel-level \vs image-level and continuous \vs categorical tasks.
    }
    
  \label{fig:uncertainty_plots}
\end{figure}

We visualize the estimated task uncertainty over the pretraining grouped for comparable targets: pixel-continuous, image-continuous, pixel-categorical and image-categorical (see Fig.~\ref{fig:uncertainty_plots}). 
For all tasks, the log variance decreases during pretraining. 
However, for the categorical tasks we observe an initial increase in uncertainty before a decrease.
Interestingly, the relative weighting of tasks changes over epochs, \eg within the pixel-continuous tasks, the Aster tasks (elevation, slope) have the highest uncertainty until epoch 150, before it drops to the lowest.
A similar change in relative weight is observed for the ERA5 and Longitude.
The latitude has a substantially lower uncertainty than the longitude, which can be explained by the rainfall and vegetation gradients along the North-South axis that shape the features and appearance of landscapes.

\begin{table}[h!]
\caption{\textbf{Loss aggregation strategy.} We evaluate the effect of the two loss aggregation strategies: Equal weighting \vs task uncertainty weighting when pretraining on MMEarth64.}
\label{tab:loss_weighting_results}
\centering
\adjustbox{max width=\textwidth}{
\begin{tabular}{@{}llcccc@{}}
\toprule
Pretrain   & Loss aggr.    & \begin{tabular}[l]{@{}c@{}}BigEarth20k\\(F1$\uparrow$) \\ FT/LP \end{tabular} & \begin{tabular}[c]{@{}c@{}}So2Sat20k\\(Acc.$\uparrow$)\\ FT/LP \end{tabular} & \begin{tabular}[c]{@{}c@{}}Cashew1k\\(IoU$\uparrow$)\\ FT\end{tabular} & \begin{tabular}[c]{@{}c@{}}SAcrop3k\\(IoU$\uparrow$)\\ FT\end{tabular} \\ \midrule
MMEarth64 & Uncertainty & 68.2/40.9 & 54.0/41.7 &      81.6 &      39.7 \\
MMEarth64 & Equal           & 68.2/36.7 & 57.0/35.9 &      81.6 &      39.3 \\
\bottomrule
\end{tabular}
}
\end{table}

\subsection{Qualitative reconstruction results}
\begin{figure}[h!]
  \centering
  \begin{subfigure}{\linewidth}
    \includegraphics[width=\linewidth]{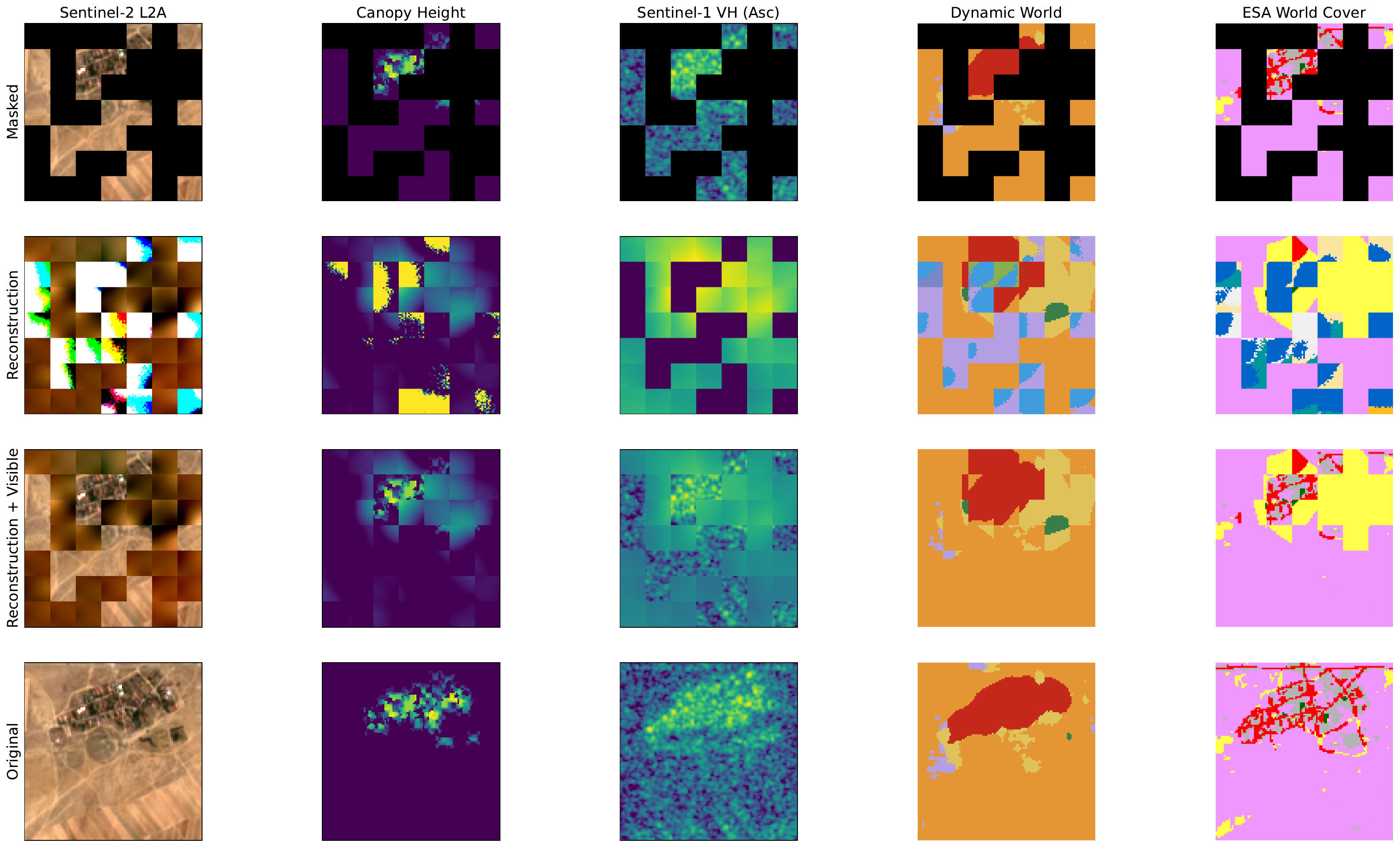}
    \caption{}
  \end{subfigure}
  \begin{subfigure}{\linewidth}
    \includegraphics[width=\linewidth]{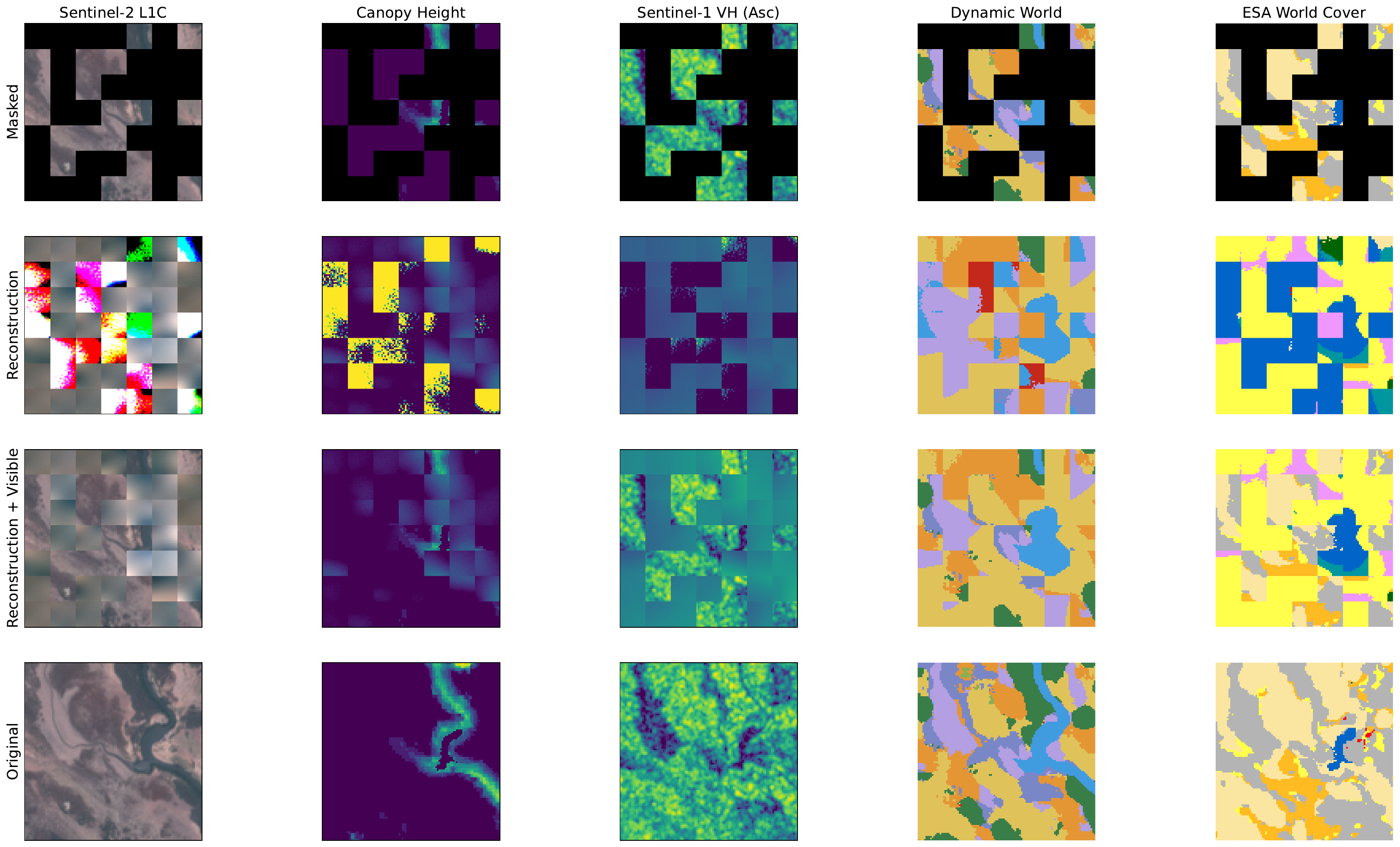}
    \caption{}
  \end{subfigure}
  
  \caption{
    \textbf{Reconstruction results.} Visualization of reconstruction examples for pixel-level pretraining tasks. 
    }
    
  \label{fig:reconstruction}
\end{figure}

We visualize some reconstruction examples of the pixel-level pretraining tasks in Fig.~\ref{fig:reconstruction}. Similar to the results in the original ConvNeXt V2 paper, we observe that the reconstruction of the masked patches is rather blurry, but coarser patterns are captured. 
The visualization of the Sentinel-2 reconstruction is also affected by the patch-level normalization of the target.
The reconstruction result of the visible patches is random, as the reconstruction loss was only optimized for non-visible patches.
However, the goal of our MP-MAE approach is not to solve the pretext tasks, but to learn good semantic representations through these tasks. 
These results show that if we were to specialize on any of the pretraining task, then one option could be to fine-tune the pretrained model separately for each task.

\end{document}